\documentclass{article}

\usepackage{microtype}
\usepackage{graphicx}
\usepackage{tabularx}
\usepackage{pbox}
\usepackage{caption} 
\usepackage{xcolor}

\usepackage{booktabs} 

\usepackage{subcaption}

\usepackage[hyphenbreaks]{breakurl}
\usepackage[hyphens]{url}
\usepackage{hyperref}

\newcolumntype{L}{>{\raggedright\arraybackslash}X}

\usepackage{amssymb}
\usepackage{bbm}
\usepackage{mathtools}

\usepackage[accepted]{include/icml2021}

\newcounter{commentCounter}
\newif\iftrvar
\trvartrue
\iftrvar
\newcommand{\tim}[1]{{\small \color{red} \refstepcounter{commentCounter}\textsf{[TR]$_{\arabic{commentCounter}}$:{#1}}}}
\newcommand{\minqi}[1]{{\small \color{blue} \refstepcounter{commentCounter}\textsf{[MJ]$_{\arabic{commentCounter}}$:{#1}}}}
\newcommand{\ed}[1]{{\small \color{magenta} \refstepcounter{commentCounter}\textsf{[ETG]$_{\arabic{commentCounter}}$:{#1}}}}

\else
\newcommand{\tim}[1]{}
\newcommand{\ed}[1]{}
\newcommand{\minqi}[1]{}

\fi

\newcommand{\algoname}{Prioritized Level Replay}
\newcommand{\algoabbrev}{PLR}

\renewcommand{\algorithmicrequire}{\textbf{Input:}}
\renewcommand{\algorithmicensure}{\textbf{Output:}}

\definecolor{dark-gray}{gray}{0.35}
\newcommand{\LineComment}[1]{{\color{dark-gray} \hfill\textit{#1}}}

\icmltitlerunning{\algoname{}}

\begin{document}

\twocolumn[
\icmltitle{\algoname{}}



\icmlsetsymbol{equal}{*}

\begin{icmlauthorlist}
\icmlauthor{Minqi Jiang}{fair,ucl}
\icmlauthor{Edward Grefenstette}{fair,ucl}
\icmlauthor{Tim Rockt\"{a}schel}{fair,ucl}
\end{icmlauthorlist}

\icmlaffiliation{fair}{Facebook AI Research, London, United Kingdom}
\icmlaffiliation{ucl}{University College London, London, United Kingdom}

\icmlcorrespondingauthor{Minqi Jiang}{msj@fb.com}

\icmlkeywords{Machine Learning, ICML, ProcGen, Active Learning}

\vskip 0.3in
]

\widowpenalty10000
\clubpenalty10000



\printAffiliationsAndNotice{}  

\begin{abstract}
Environments with procedurally generated content serve as important benchmarks for testing systematic generalization in deep reinforcement learning. In this setting, each \emph{level} is an algorithmically created environment instance with a unique configuration of its factors of variation.
Training on a prespecified subset of levels allows for testing generalization to unseen levels. What can be learned from a level depends on the current policy, yet prior work defaults to uniform sampling of training levels independently of the policy. We introduce \emph{Prioritized Level Replay} (PLR), a general framework for selectively sampling the next training level by prioritizing those with higher estimated learning potential when revisited in the future. We show TD-errors effectively estimate a level's future learning potential and, when used to guide the sampling procedure, induce an emergent curriculum of increasingly difficult levels. By adapting the sampling of training levels, PLR significantly improves sample efficiency and generalization on Procgen Benchmark—matching the previous state-of-the-art in test return—and readily combines with other methods. Combined with the previous leading method, PLR raises the state-of-the-art to over 76\% improvement in test return relative to standard RL baselines. 

\end{abstract}

\section{Introduction}

\begin{figure*}[t!]
    \centering
    \includegraphics[width=1.0\linewidth]{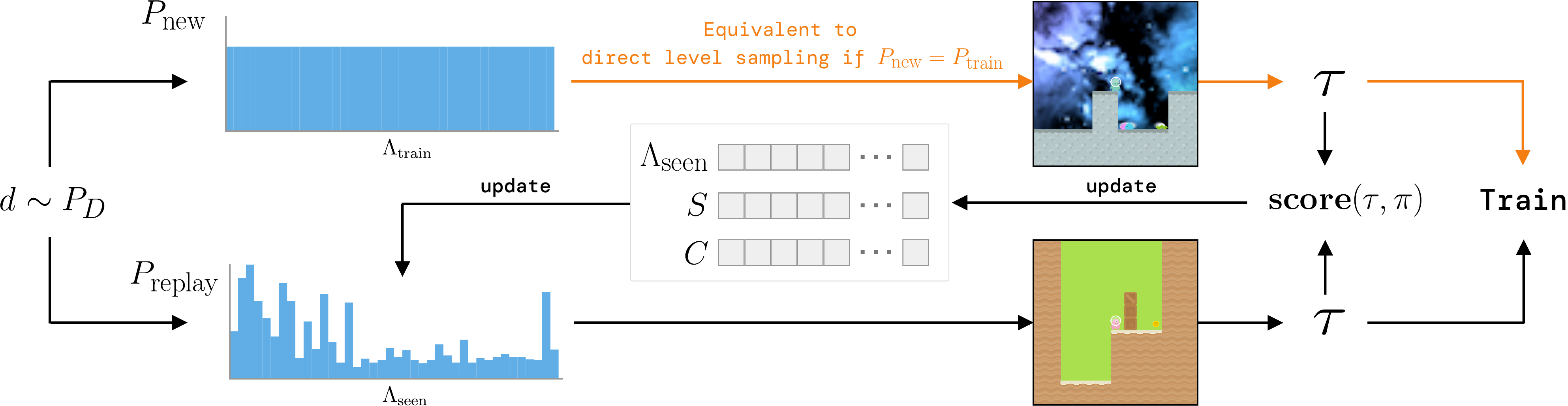}
    \caption{Overview of \algoname{}. The next level is either sampled from a distribution with support over unseen levels (top), which could be the environment's (perhaps implicit) full training-level distribution, or alternatively, sampled from the replay distribution, which prioritizes levels based on future learning potential (bottom). In either case, a trajectory $\tau$ is sampled from the next level and used to update the replay distribution. This update depends on the lists of previously seen levels  $\Lambda_{\text{seen}}$, their latest estimated learning potentials $S$, and last sampled timestamps $C$.}
    \label{fig:algo_fig}
\end{figure*}

Deep reinforcement learning (RL) easily overfits to training experiences, making generalization a key open challenge to widespread deployment of these methods. Procedural content generation (PCG) environments have emerged as a promising class of problems with which to probe and address this core weakness \citep{Risi_2020, gym_minigrid, cobbe2019procgen, Juliani_2019, DBLP:conf/iclr/ZhongRG20, kttler2020nethack}. Unlike singleton environments, like the Arcade Learning Environment games \citep{Bellemare_2013}, PCG environments take on algorithmically created configurations at the start of each training episode, potentially varying the layout, asset appearances, or even game dynamics. Each environment instance generated this way is called a \emph{level}. 
In mapping an identifier, such as a random seed, to each level, PCG environments allow us to measure a policy's generalization to held-out test levels. In this paper, we assume only a blackbox generation process that returns a level given an identifier. We avoid the strong assumption of control over the generation procedure itself, explored by several prior works (see Section \ref{section:related_work}). Further, we assume levels follow common latent dynamics, so that in aggregate, experiences collected in individual levels reveal general rules governing the entire set of levels.

Despite its humble origin in games, the PCG abstraction proves far-reaching: Many control problems, such as teaching a robotic arm to stack blocks in a specific formation, easily conform to PCG. Here, each level may consist of a unique combination of initial block positions, arm state, and target formation. In a vastly different domain, Hanabi \citep{bard2020hanabi} also conforms to PCG, where levels map to initial deck orderings. These examples illustrate the generality of the PCG abstraction: Most challenging RL problems entail generalizing across instances (or levels) differing along some underlying factors of variation and thereby can be aptly framed as PCG. This highlights the importance of effective deep RL methods for PCG environments.

Many techniques have been proposed to improve generalization in the PCG setting (see Section \ref{section:related_work}), requiring changes to the model architecture, learning algorithm, observation space, or environment structure. Notably, these approaches default to uniform sampling of training levels. We instead hypothesize that the variation across levels implies that at each point of training, each level likely holds different potential for an agent to learn about the structure shared across levels to improve generalization. Inspired by this insight and selective sampling methods from active learning, we investigate whether sampling the next training level weighed by its learning potential can improve generalization.

We introduce \algoname{} (PLR), illustrated in Figure~\ref{fig:algo_fig}, a method for sampling training levels that exploits the differences in learning potential among levels to improve both sample efficiency and generalization. \algoabbrev{}  selectively samples the next training level based on an estimated learning potential of replaying each level anew. 
During training, our method updates scores estimating each level's learning potential as a function of the agent's policy and temporal-difference (TD) errors collected along the last trajectory sampled on that level. 
Our method then samples the next training level from a distribution derived from a normalization procedure over these level scores. \algoabbrev{} makes no assumptions about how the policy is updated, so it can be used in tandem with any RL algorithm and combined with many other methods such as data augmentation. 
Our method also does not assume any external, predefined ordering of levels by difficulty or other criteria, but instead derives level scores dynamically during training based on how the policy interacts with the environment. The only requirements are as follows---satisfied by almost any problem that can be framed as PCG, including RL environments implemented as seeded simulators: (i) Some notion of ``level'' exists, such that levels follow common latent dynamics; (ii) such levels can be sampled from the environment in an identifiable way; and (iii) given a level identifier, the environment can be set to that level to collect new experiences from it.

While previous works in off-policy RL devised effective methods to directly reuse \emph{past} experiences for training \citep{schaul2015prioritized, andrychowicz2017hindsight}, \algoabbrev{} uses past experiences to inform the collection of \emph{future} experiences by estimating how much replaying each level anew will benefit learning. Our method can thus be seen as a forward-view variation of prioritized experience replay, and an online counterpart to this off-policy method.

This paper makes the following contributions\footnote{Our code is available at \url{https://github.com/facebookresearch/level-replay}.}: 
(i) We introduce a computationally-efficient algorithm for sampling levels during training based on an estimate of the future learning potential of collecting new experiences from each level;
(ii) we show our method significantly improves generalization on 10 of 16 environments in Procgen Benchmark and two challenging MiniGrid environments; (iii) we demonstrate our method combines with the previous leading method to set a new state-of-the-art on Procgen Benchmark; and (iv) we show our method induces an implicit curriculum over training levels in sparse reward settings.
\section{Background}
In this paper, we refer to a \emph{PCG environment} as any computational process that, given a level identifier (e.g.~an index or a random seed), generates a \emph{level}, defined as an environment instance exhibiting a unique configuration of its underlying factors of variation, such as layout, asset appearances, or specific environment dynamics~\cite{Risi_2020}. For example, MiniGrid's MultiRoom environment instantiates mazes with varying numbers of rooms based on the seed \citep{gym_minigrid}. We refer to sampling a new trajectory generated from the agent's latest policy acting on a given level $l$ as \emph{replaying} that level $l$.

The level diversity of PCG environments makes them useful testbeds for studying the robustness and generalization ability of RL agents, measured by agent performance on unseen test levels. The standard test evaluation protocol for PCG environments consists of training the agent on a finite number of training levels $\Lambda_{\text{train}}$, and evaluating performance on unseen test levels $\Lambda_{\text{test}}$, drawn from the set of all levels. Training levels are sampled from an arbitrary distribution $P_{\text{train}}(l|\Lambda_{\text{train}})$. We call this training process \emph{direct level sampling}. A common variation of this protocol sets $\Lambda_{\text{train}}$ to the set of all levels, though in practice, the agent will still only effectively see a finite set of levels after training for a finite number of steps. In the case of a finite training set, typically $P_{\text{train}}(l|\Lambda_{\text{train}}) = \mathbf{Uniform}(l; \Lambda_{\text{train}})$. See Appendix~\ref{app:algorithms} for the pseudocode outlining this procedure.

PCG environments naturally lend themselves to curriculum learning. Prior works have shown that directly altering levels to match their difficulty to the agent's abilities can improve generalization \citep{justesen2018illuminating, dennis2020emergent,DBLP:journals/corr/abs-1810-08272,DBLP:conf/iclr/ZhongRG20}. These findings further suggest the levels most useful for improving an agent's policy vary throughout the course of training. In this work, we consider how to automatically discover a curriculum that improves generalization for a general blackbox PCG environment---crucially, without assuming any knowledge or control of how levels are generated (beyond providing the random seed or other indicial level identifier).
\section{\algoname{}}
\label{section:methods}
In this section, we present~\emph{\algoname{}}~(\algoabbrev{}), an algorithm for selectively sampling the next training level given the current policy, by prioritizing levels with higher estimated learning potential when replayed (that is, revisited). 
\algoabbrev{} is a drop-in replacement for the experience-collection process used in a wide range of RL algorithms. Algorithm~\ref{alg:pg_level_replay} shows how it is straightforward to incorporate \algoabbrev{} into a generic policy-gradient training loop. For clarity, we focus on training on batches of complete trajectories (see Appendix~\ref{app:algorithms} for pseudocode of \algoabbrev{} with $T$-step rollouts).

Our method, illustrated in Figure~\ref{fig:algo_fig} and fully specified in Algorithm~\ref{alg:level_replay_mc}, induces a dynamic, nonparametric sampling distribution $P_{\text{replay}}(l | \Lambda_{\text{seen}})$ over previously visited training levels $\Lambda_{\text{seen}}$ that prioritizes visited levels with higher learning potential based on properties of the agent's past trajectories. We refer to $P_{\text{replay}}(l | \Lambda_{\text{seen}})$ as the \emph{replay distribution}. 
Throughout training, our method updates this replay distribution according to a heuristic score, assigning greater weight to visited levels with higher \emph{future} learning potential. 
Using dynamic arrays $S$ and $C$ of equal length to $\Lambda_{\text{seen}}$, \algoabbrev{} tracks level scores $S_i \in S$ for each visited training level $l_i$ based on the latest episode trajectory on $l_i$, as well as the episode count $C_i \in C$ at which each level $l_i \in \Lambda_{\text{seen}}$ was last sampled. Our method updates $P_{\text{replay}}$ after each terminated episode by computing a mixture of two distributions, $P_S$, based on the level scores, and $P_C$, based on how long ago each level was last sampled:
\begin{equation}
\label{eq:replay}
\begin{aligned}
P_{\text{replay}} = (1-\rho) \cdot P_S + \rho \cdot P_C,
\end{aligned}
\end{equation}
where the staleness coefficient $\rho \in [0,1]$ is a hyperparameter. 
We discuss how we compute level scores $S_i$, parameterizing the scoring distribution $P_S$, and the staleness distribution $P_C$, in Sections~\ref{sec:scoring} and~\ref{sec:staleness}, respectively.

\algoabbrev{} chooses the next level at the start of every training episode by first sampling a replay-decision from a Bernoulli (or similar) distribution $P_D(d)$ to determine whether to replay a level sampled from the replay distribution $P_{\text{replay}}(l|\Lambda_{\text{seen}})$ or to experience a new, unseen level from $\Lambda_{\text{train}}$, according to some distribution $P_{\text{new}}(l|\Lambda_{\text{train}} \setminus \Lambda_{\text{seen}})$. In practice, for the case of a finite number of training levels, we implement $P_{\text{new}}$ as a uniform distribution over the remaining unseen levels. For the case of a countably infinite number of training levels, we simulate $P_{\text{new}}$ by sampling levels from $P_{\text{train}}$ until encountering an unseen level. 
In our experiments based on a finite number of training levels, we opt to naturally anneal $P_D(d=1)$ as $|\Lambda_{\text{seen}}|/|\Lambda_{\text{train}}|$, so replay occurs more often as more training levels are visited.

The following sections describes how \algoname{} updates the replay distribution $P_{\text{replay}}(l | \Lambda_{\text{seen}})$, namely through level scoring and staleness-aware prioritization.
 
\begin{figure}[t!] 
\vskip -0.1in
\begin{minipage}{\linewidth}
\begin{algorithm}[H]
{
\caption{Policy-gradient training loop with \algoabbrev{}}
\label{alg:pg_level_replay}
\small
\begin{algorithmic}
    \REQUIRE Training levels $\Lambda_{\text{train}}$, policy $\pi_{\theta}$, \\ policy update function $\mathcal{U}(\mathcal{B}, \theta) \rightarrow \theta'$, and batch size $N_{b}$.
    \STATE Initialize level scores $S$ and level timestamps $C$
    \STATE Initialize global episode counter $c \gets 0$
    \STATE Initialize the ordered set of visited levels $\Lambda_{\text{seen}} = \varnothing$ 
    \STATE Initialize experience buffer $\mathcal{B} = \varnothing$
    \WHILE{training}
        \STATE $\mathcal{B} \leftarrow \varnothing$
        \WHILE{collecting experiences}
            \STATE $\mathcal{B} \leftarrow \mathcal{B}\;\cup\;\textbf{collect\_experiences}( \Lambda_{\text{train}}, \Lambda_{\text{seen}}, \pi_{\theta}, S, C, c)$ \\ \LineComment{Using Algorithm~\ref{alg:level_replay_mc}}
        \ENDWHILE
        \STATE{$\theta \leftarrow \mathcal{U}(\mathcal{B}, \theta)$ 
        \LineComment{Update policy using collected experiences}} 
    \ENDWHILE
\end{algorithmic}
}
\end{algorithm}
\vspace{-.5cm}
\begin{algorithm}[H]
\caption{Experience collection with PLR}
\label{alg:level_replay_mc}
{
\small
\begin{algorithmic}
    \REQUIRE Training levels $\Lambda_{\text{train}}$, visited levels $\Lambda_{\text{seen}}$, policy $\pi$,\\ global level scores $S$, global level timestamps $C$, and\\ global episode counter $c$.
    \ENSURE A sampled trajectory $\tau$
        \STATE $c \gets c + 1$
        \STATE Sample replay decision $d \sim{} P_D(d)$
        \IF{$d = 0$ \textbf{and} $\left|\Lambda_{\text{train}} \setminus \Lambda_{\text{seen}}\right| > 0$} 
            \STATE Define new index $i \gets |S| + 1$
            \STATE Sample $l_i \sim{} P_{\text{new}}(l | \Lambda_{\text{train}}, \Lambda_{\text{seen}})$ \LineComment{Sample an unseen level}
            \STATE Add $l_i$ to $\Lambda_{\text{seen}}$
            \STATE Add initial value $S_i = 0$ to $S$ and $C_i = 0$ to $C$
        \ELSE 
            \STATE Sample $l_i \sim{} (1-\rho) \cdot P_S(l | \Lambda_{\text{seen}}, S) + \rho \cdot P_C(l | \Lambda_{\text{seen}}, C, c)$
            \STATE \LineComment{Sample a level for replay}
        \ENDIF
        
    
        \STATE Sample $\tau \sim{} P_\pi(\tau | l_i)$
        
        \STATE Update score $S_i \gets \mathbf{score}(\tau, \pi)$ and timestamp $C_i \gets c$ 
        

\end{algorithmic}
}
\end{algorithm}
\vspace{-1cm}
\end{minipage}
\end{figure}

\subsection{Scoring Levels for Learning Potential}
\label{sec:scoring}
After collecting each complete episode trajectory $\tau$ on level $l_i$ using policy $\pi$, our method assigns $l_i$ a score \mbox{$S_i = \mathbf{score}(\tau, \pi)$} measuring the learning potential of replaying $l_i$ in the future. We employ a function of the TD-error at timestep $t$, $\delta_t = r_t + \gamma V(s_{t+1}) - V(s_t)$, as a proxy for this learning potential. The expectation of the TD-error over next states is equivalent to the advantage estimate, and therefore higher-magnitude TD-errors imply greater discrepancy between expected and actual returns, making $\delta_t$ a useful measure of the learning potential in revisiting a particular state transition. To prioritize the learning potential of future experiences resulting from replaying a level, we use a scoring function based on the \emph{average magnitude} of the Generalized Advantage Estimate~\citep[GAE;][]{schulman2018high} over each of $T$ time steps in the latest trajectory $\tau$ from that level:

\vspace{-0.4cm}
\begin{equation}
S_i = \mathbf{score}(\tau, \pi) = \frac{1}{T}\sum_{t=0}^{T} \left|\sum_{k=t}^T(\gamma\lambda)^{k-t}\delta_k\right|. \label{eq:average_scoring_function}
\vspace{-0.2cm}
\end{equation}


While the GAE at time $t$ is most commonly expressed as the discounted sum of all $1$-step TD-errors starting at $t$ as in Equation~\ref{eq:average_scoring_function}, it is equivalent to an exponentially-discounted sum of all $k$-step TD-errors from $t$, with discount factor $\lambda$. By considering all $k$-step TD-errors, the GAE mitigates the bias introduced by the bootstrap term in $1$-step TD-errors. The discount factor $\lambda$ then controls the trade-off between bias and variance. Our scoring function considers the absolute value of the GAE, as we assume the learning potential grows with the magnitude of the TD-error irrespective of its sign. This also avoids opposite signed errors canceling out.


Another useful interpretation of Equation~\ref{eq:average_scoring_function} comes from observing that the GAE magnitude at $t$ is equivalent to the L1 value loss $|\hat{V}_t - V_t|$ under a policy-gradient algorithm that uses GAE for its own advantage estimates (and therefore value targets $\hat{V}_t$), as done in state-of-the-art implementations of PPO \citep{DBLP:journals/corr/SchulmanWDRK17} used in our experiments. 
Unless otherwise indicated, \algoabbrev{} refers to the instantiation of our algorithm with L1 value loss as the scoring function.

We further formulate the \emph{Value Correction Hypothesis} to motivate our approach: In sparse reward settings, prioritizing the sampling of training levels with greatest average absolute value loss leads to a curriculum that improves both sample efficiency and generalization. We reason that on threshold levels (i.e.~those at the limit of the agent's current abilities) the agent will see non-stationary returns (or value targets)---and therefore incur relatively high value errors---until it learns to solve them consistently.
In contrast, levels beyond the agent's current abilities tend to result in stationary value targets signaling failure and therefore low value errors, until the agent learns useful, transferable behaviors from threshold levels.
Prioritizing levels by value loss then naturally guides the agent along the expanding threshold of its ability---without the need for any externally provided measure of difficulty. We believe that learning behaviors systematically aligned with the inherent complexities of the environment in this way may lead to better generalization, and will seek to verify this empirically in Section~\ref{subsection:minigrid_results}.

While we provide principled motivations for our specific choice of scoring function, we emphasize that in general, the scoring function can be any approximation of learning potential based on trajectory values. Note that candidate scoring functions should asymptotically decrease with frequency of level visitation to avoid mode collapse of $P_{\text{replay}}$ to a limited set of levels and possible overfitting. In Section~\ref{sec:experiments}, we compare our choice of the GAE magnitude, or equivalently, the L1 value loss, to alternative TD-error-based and uncertainty-based scoring approaches, listed in Table~\ref{table:scoring_metrics}.

Given level scores, we use normalized outputs of a prioritization function $h$ evaluated over these scores and tuned using a temperature parameter $\beta$ to define the score-prioritized distribution $P_S(\Lambda_{\text{train}})$ over the training levels, under which
\begin{equation}
\label{eq:score}
\begin{aligned}
P_S(l_i|\Lambda_{\text{seen}}, S) = \frac{h(S_i)^{1/\beta}}{\sum_j h(S_j)^{1/\beta}}. 
\end{aligned}
\end{equation}
The function $h$ defines how differences in level scores translate into differences in prioritization. The temperature parameter $\beta$ allows us to  tune how much $h(S)$ ultimately determines the resulting distribution. We make the design choice of using rank prioritization, for which $h(S_i) = 1/\text{rank}(S_i)$, where $\text{rank}(S_i)$ is the rank of level score $S_i$ among all scores sorted in descending order. We also experimented with proportional prioritization ($h(S_i)=S_i$) as well as greedy prioritization (the level with the highest score receives probability~$1$), both of which tend to perform worse.

\subsection{Staleness-Aware Prioritization}
\label{sec:staleness}
As the scores used to parameterize $P_S$ are a function of the state of the policy at the time the associated level was last played, they come to reflect a gradually more off-policy measure the longer they remain without an update through replay. We mitigate this drift towards ``off-policy-ness'' by explicitly mixing the sampling distribution with a staleness-prioritized distribution $P_C$:
\begin{equation}
\label{eq:staleness}
\begin{aligned}
P_C(l_i | \Lambda_{\text{seen}}, C, c) = \frac{c - C_i}{\sum_{C_j \in C} c - C_j}
\end{aligned}
\end{equation}
which assigns probability mass to each level $l_i$ in proportion to the level's \emph{staleness} $c - C_i$. 
Here, $c$ is the count of total episodes sampled so far in training and $C_i$ (referred to as the level's timestamp) is the episode count at which $l_i$ was last sampled. By pushing support to levels with staler scores, $P_C$ ensures no score drifts too far off-policy. 

Plugging Equations~\ref{eq:score} and~\ref{eq:staleness} into Equation~\ref{eq:replay} gives us a replay distribution that is calculated as
\[
P_{\text{replay}}(l_i) = (1-\rho) \cdot P_S(l_i | \Lambda_{\text{seen}}, S) + \rho \cdot P_C(l_i | \Lambda_{\text{seen}}, C, c).
\]
Thus, a level has a greater chance of being sampled when its score is high or it has not been sampled for a long time.

\section{Experimental Setting}
\label{sec:experiments}
We evaluate \algoabbrev{} on several PCG environments with various combinations of scoring functions and prioritization schemes, and compare to the most common direct level sampling baseline of $P_{\text{train}}(l | \Lambda_{\text{train}}) = \mathbf{Uniform}(l; \Lambda_{\text{train}})$. We train and test on all 16 environments in the Procgen Benchmark on easy and hard difficulties, but focus discussion on the easy results, which allow direct comparison to several prior studies. We compare to UCB-DrAC~\citep{raileanu2021automatic}, the state-of-the-art image augmentation method on this benchmark, and mixreg~\citep{wang2020mixreg}, a recently introduced data augmentation method. We also compare to TSCL Window~\citep{Matiisen_2020}, which resembles \algoabbrev{} with an alternative scoring function using the slope of recent returns and no staleness sampling. For fair comparison, we also evaluate a custom TSCL Window variant that mixes in the staleness distribution $P_{C}$ weighted by $\rho > 0$. Further, to demonstrate the ease of combining \algoabbrev{} with other methods, we evaluate UCB-DrAC using PLR for sampling training levels. Finally, we test the Value Correction Hypothesis on two challenging MiniGrid environments.

We measure episodic \emph{test returns} per game throughout training, as well as the performance of the final policy over 100 unseen test levels of each game relative to PPO with uniform sampling. We also evaluate the mean normalized episodic test return and mean generalization gap, averaged over all games (10 runs each). We normalize returns according to \citet{cobbe2019quantifying} and compute the generalization gap as train returns minus test returns. Thus, a larger gap indicates more overfitting, making it an apt measure of generalization. We assess statistical significance at $p=0.05$, using the Welch t-test.

In line with the standard baseline for these environments, all experiments use PPO with GAE for training. 
For \mbox{Procgen}, we use the same ResBlock architecture as \citet{cobbe2019procgen} and train for 25M total steps on 200 levels on the easy setting as in the original baselines. 
For MiniGrid, we use a 3-layer CNN architecture based on \citet{igl2019generalization}, and provide approximately 1000 levels of each difficulty per environment during training. Detailed descriptions of the environments, architectures, and hyperparameters used in our experiments (and how they were set or obtained) can be found in Appendix~\ref{app:environments}. See Table~\ref{table:scoring_metrics} for the full set of scoring functions investigated in our experiments. 

Additionally, we extend PLR to support training on an unbounded number of levels by tracking a rolling, finite buffer of the top levels so far encountered by learning potential. Appendix \ref{section:training_on_full_dist} reports the results of this extended PLR \mbox{algorithm} when training on the full level distribution of the MiniGrid environments featured in our main experiments.

\begingroup
\setlength{\tabcolsep}{1pt}
\begin{table}[hbtp]
\vskip -0.1in
\caption{Scoring functions investigated in this work.}
\vskip 0.1in
\label{table:scoring_metrics}
\begin{center}
\begin{tabularx}{\linewidth}{p{0.42\linewidth}p{0.5\linewidth}} 
\toprule
\renewcommand{\arraystretch}{1.8}
\textbf{Scoring function} & \textbf{$\mathbf{score}(\tau, \pi)$} \\ 
\midrule
Policy entropy & $\frac{1}{T}\sum_{t=0}^{T} \sum_a{\pi(a, s_t) \log \pi(a, s_t)}$ \\[8pt]
Policy min-margin & \pbox{7cm}{$\frac{1}{T}\sum_{t=0}^{T} (\max_{a} \pi(a, s_t$) $-$\\ $\max_{a \neq \max_{a} \pi(a, s_t)} \pi(a, s_t))$} \\[8pt]
Policy least-confidence & $\frac{1}{T}\sum_{t=0}^{T} (1 - \max_{a}\pi(a, s_t))$ \\[8pt]
1-step TD error & $\frac{1}{T}\sum_{t=0}^{T} |\delta_t|$ \\[8pt]
GAE & $\frac{1}{T}\sum_{t=0}^{T} \sum_{k=t}^T(\gamma\lambda)^{k-t}\delta_k$ \\[8pt]
\pbox{4cm}{GAE magnitude \\ (L1 value loss)}  & $\frac{1}{T}\sum_{t=0}^{T} \left|\sum_{k=t}^T(\gamma\lambda)^{k-t}\delta_k\right|$ \\ 
\bottomrule
\end{tabularx}
\end{center}
\vskip -0.15in
\end{table}
\endgroup
\section{Results and Discussion}

\begin{figure*}[t!]
    \centering
    \includegraphics[width=1.0\textwidth]{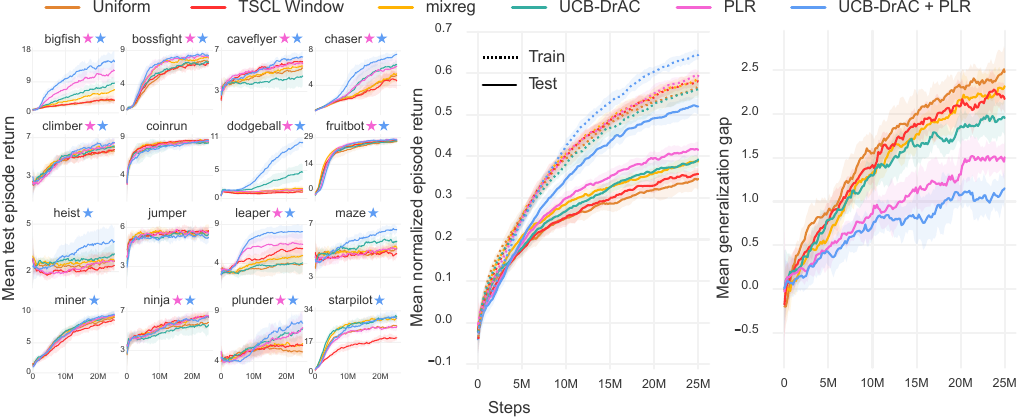}
    \caption{Left: Mean episodic test returns (10 runs) of each method. Each colored $\bigstar$ indicates statistically significant ($p<0.05$) gains in final test performance or sample complexity along the curve, relative to uniform sampling, for the PLR-based method of the same color. Center: Mean normalized train and test returns averaged across all games. Right: Mean generalization gaps averaged across all games.}
    \label{fig:procgen_summary}
\end{figure*}

Our main findings are that
(i) \algoabbrev{} significantly improves both sample efficiency and generalization, attaining the highest normalized mean test and train returns and mean reduction in generalization gap on Procgen out of all individual methods evaluated, while matching UCB-DrAC in test improvement relative to PPO;
(ii) alternative scoring functions lead to inconsistent improvements across environments; (iii) \algoabbrev{} combined with UCB-DrAC sets a new state-of-the-art on Procgen; and (iv) \algoabbrev{} induces an implicit curriculum over training levels, which substantially aids training in two challenging MiniGrid environments.

\subsection{Procgen Benchmark}
\label{subsection:procgen_benchmark}
\begin{figure}[t!]
    \centering
    \includegraphics[width=1.0\linewidth]{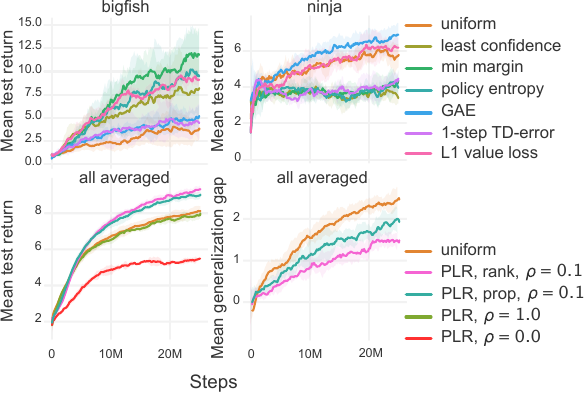}
    \vspace{-.2cm}
    \caption{Top: Two example Procgen environments, between which all scoring functions except L1 value loss show inconsistent improvements to test performance (rank prioritization, \mbox{$\beta=0.1$}, $\rho=0.3$). This inconsistency holds across settings in our grid search. Bottom: Mean unnormalized episodic test returns (left) and mean generalization gap (right) for various PLR settings.}
    \label{fig:alternative-settings-results}
    \vskip -.2in
\end{figure}

Our results, summarized in Figure~\ref{fig:procgen_summary}, show \algoabbrev{} with rank prioritization ($\beta = 0.1$, $\rho = 0.1$) leads to the largest statistically significant gains in mean normalized test and train returns and reduction in generalization gap compared to uniform sampling, outperforming all other methods besides UCB-DrAC + \algoabbrev{}. \algoabbrev{} combined with UCB-DrAC sees the most drastic improvements in these metrics. As reported in Table \ref{tab:ucb-drac-plr-sota}, UCB-DrAC + \algoabbrev{} yields a 76\% improvement in mean test return relative to PPO with uniform sampling, and a 28\% improvement relative to the previous state-of-the-art set by UCB-DrAC. While \algoabbrev{} with rank prioritization leads to statistically significant gains in test return on 10 of 16 environments and proportional prioritization, on 11 of 16 games, we prefer rank prioritization: While we find the two comparable in mean normalized returns, Figure \ref{fig:alternative-settings-results} shows rank prioritization results in higher mean \emph{unnormalized} test returns and a significantly lower mean generalization gap, averaged over all environments.

Further, Figure \ref{fig:alternative-settings-results} shows that gains only occur when $P_{\text{replay}}$ considers \emph{both} level scores and staleness ($0 < \rho < 1$), highlighting the importance of staleness-based sampling in keeping scores from drifting off-policy. Lastly, we also benchmarked \algoabbrev{} on the hard setting against the same set of methods, where it again leads with 35\% greater test returns relative to uniform sampling and 83\% greater test returns when combined with UCB-DrAC. Figures
\mbox{\ref{fig:procgen_hard_summary}--\ref{fig:rank_prop_train_eval}}
and Table \ref{tab:ucb-drac-plr-hard} in Appendix~\ref{app:additional_experiments} report additional details on these results.

The alternative scoring metrics based on TD-error and classifier uncertainty perform inconsistently across games. While certain games, such as BigFish, see improved sample-efficiency and generalization under various scoring functions, others, such as Ninja, see no improvement or worse, degraded performance. See Figure~\ref{fig:alternative-settings-results} for an example of this inconsistent effect across games. We find the best-performing variant of TSCL Window does not incorporate staleness information ($\rho = 0$) and similarly leads to inconsistent outcomes across games at test time, notably significantly worsening performance on StarPilot, as seen in Figure \ref{fig:procgen_summary}, and increasing the generalization gap on some environments as revealed in Figure \ref{fig:procgen_gen_gap_large} of Appendix \ref{app:additional_experiments}.

\begin{table*}[t!]
\vskip -0.1in
\newcommand\mc[1]{\multicolumn{1}{l}{#1}}

\setlength{\tabcolsep}{2.5pt}
\caption{Test returns of policies trained using each method with its best hyperparameters. Following \citet{raileanu2021automatic}, the reported mean and standard deviations per environment are computed by evaluating the final policy's average return on 100 test episodes, aggregated across multiple training runs (10 runs for Procgen Benchmark and 3 for MiniGrid, each initialized with a different training seed). Normalized test returns per run are computed by dividing the average test return per run for each environment by the corresponding average test return of the uniform-sampling baseline over all runs. We then report the means and standard deviations of normalized test returns aggregated across runs. We report the normalized return statistics for Procgen and MiniGrid environments separately. Bolded methods are not significantly different from the method with highest mean, unless all are, in which case none are bolded.}
\label{tab:ucb-drac-plr-sota}

\begin{center}
\begin{subtable}{\linewidth}
\begin{tabular}{l|r r r r| r r}
\toprule
Environment &Uniform &TSCL &mixreg &UCB-DrAC &PLR &UCB-DrAC + PLR\\
\midrule
BigFish&$3.7\pm1.2$&$4.3\pm1.3$&$6.9\pm1.6$&$8.7\pm1.1$&$10.9\pm2.8$&$\mathbf{14.3\pm2.1}$\\
BossFight&$7.7\pm0.4$&$7.4\pm0.8$&$8.1\pm0.7$&$7.7\pm0.7$&$\mathbf{8.9\pm0.4}$&$\mathbf{8.8\pm0.8}$\\
CaveFlyer&$5.4\pm0.8$&$\mathbf{6.3\pm0.6}$&$6.0\pm0.6$&$4.6\pm0.9$&$\mathbf{6.3\pm0.5}$&$\mathbf{6.8\pm0.7}$\\
Chaser&$5.2\pm0.7$&$4.9\pm1.0$&$5.7\pm1.1$&$6.8\pm0.9$&$6.9\pm1.2$&$\mathbf{8.0\pm0.6}$\\
Climber&$5.9\pm0.6$&$6.0\pm0.8$&$\mathbf{6.6\pm0.7}$&$\mathbf{6.4\pm0.9}$&$\mathbf{6.3\pm0.8}$&$\mathbf{6.8\pm0.7}$\\
CoinRun&$8.6\pm0.4$&$\mathbf{9.2\pm0.2}$&$8.6\pm0.3$&$8.6\pm0.4$&$8.8\pm0.5$&$\mathbf{9.0\pm0.4}$\\
Dodgeball&$1.7\pm0.2$&$1.2\pm0.4$&$1.8\pm0.4$&$5.1\pm1.6$&$1.8\pm0.5$&$\mathbf{10.3\pm1.4}$\\
FruitBot&$27.3\pm0.9$&$27.1\pm1.6$&$27.7\pm0.8$&$27.0\pm1.3$&$28.0\pm1.3$&$27.6\pm1.5$\\
Heist&$2.8\pm0.9$&$2.5\pm0.6$&$2.7\pm0.4$&$3.2\pm0.7$&$2.9\pm0.5$&$\mathbf{4.9\pm1.3}$\\
Jumper&$5.7\pm0.4$&$6.1\pm0.6$&$6.1\pm0.3$&$5.6\pm0.5$&$5.8\pm0.5$&$5.9\pm0.3$\\
Leaper&$4.2\pm1.3$&$6.4\pm1.2$&$5.2\pm1.1$&$4.4\pm1.4$&$6.8\pm1.2$&$\mathbf{8.7\pm1.0}$\\
Maze&$5.5\pm0.4$&$5.0\pm0.3$&$5.4\pm0.5$&$6.2\pm0.5$&$5.5\pm0.8$&$\mathbf{7.2\pm0.8}$\\
Miner&$8.7\pm0.7$&$8.9\pm0.6$&$9.5\pm0.4$&$\mathbf{10.1\pm0.6}$&$\mathbf{9.6\pm0.6}$&$\mathbf{10.0\pm0.5}$\\
Ninja&$6.0\pm0.4$&$\mathbf{6.8\pm0.5}$&$\mathbf{6.9\pm0.5}$&$5.8\pm0.8$&$\mathbf{7.2\pm0.4}$&$\mathbf{7.0\pm0.5}$\\
Plunder&$5.1\pm0.6$&$5.9\pm1.1$&$5.7\pm0.5$&$\mathbf{7.8\pm0.9}$&$\mathbf{8.7\pm2.2}$&$\mathbf{7.7\pm0.9}$\\
StarPilot&$26.8\pm1.5$&$19.8\pm3.4$&$\mathbf{32.7\pm1.5}$&$\mathbf{31.7\pm2.4}$&$27.9\pm4.4$&$29.6\pm2.2$\\
\midrule
Normalized test returns (\%)&$100.0\pm4.5$&$103.0\pm3.6$&$113.8\pm2.8$&$129.8\pm8.2$&$128.3\pm5.8$&$\mathbf{176.4\pm6.1}$\\
\midrule
\addlinespace[0.25cm]
MultiRoom-N4-Random & $0.80\pm0.04$ & -- & -- & -- & $\mathbf{0.81\pm0.01}$ & -- \\
ObstructedMazeGamut-Easy&$0.53\pm0.04$& -- & -- & -- & $\mathbf{0.85\pm0.04}$& -- \\
ObstructedMazeGamut-Med&$0.65\pm0.01$& -- & -- & -- & $\mathbf{0.73\pm0.07}$& -- \\
\midrule
Normalized test returns (\%)&$100.0\pm2.5$& -- & -- & -- &$\mathbf{124.3\pm4.7}$ & --\\
\bottomrule
\end{tabular}
\end{subtable}
\end{center}
\end{table*}

\subsection{MiniGrid}
\label{subsection:minigrid_results}
\begin{figure*}
    \centering
    \begin{subfigure}[b]{.9\textwidth}
        \includegraphics[width=\textwidth]{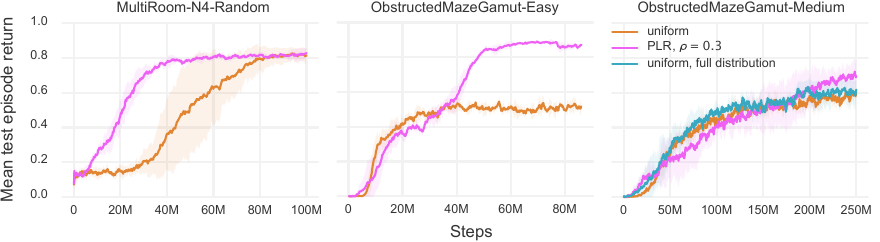}
        \label{fig:multiroom-value-testreturns}
    \end{subfigure}\vspace{-1em}
    \begin{subfigure}[b]{0.9\textwidth}
        \includegraphics[width=\textwidth]{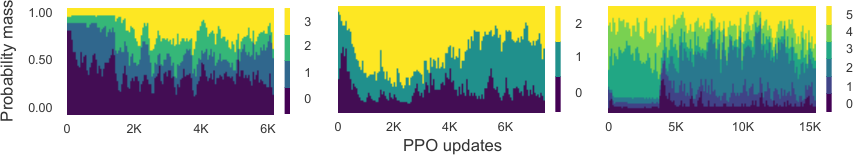}
        \label{fig:minigrid-autocurricula-final}
    \end{subfigure}
    \vspace{-.5cm}
    \caption{Top: Mean episodic test returns of \algoabbrev{} and the uniform-sampling baseline on MultiRoom-N4-Random (4 runs), ObstructedMazeGamut-Easy (3 runs), and ObstructedMazeGamut-Medium (3 runs). Bottom: The probability mass assigned to levels of varying difficulty over the course of training in a single, randomly selected run for the respective environment.
    }
    \label{fig:minigrid-results}
\end{figure*}

We provide empirical support for the Value Correction Hypothesis (defined in Section \ref{section:methods}) on two challenging MiniGrid environments, whose levels fall into discrete difficulties (e.g. by number of rooms to be traversed). In both, \algoabbrev{} with rank prioritization significantly improves sample efficiency and generalization over uniform sampling, demonstrating our method also works well in discrete state spaces. We find a staleness coefficient of $\rho = 0.3$ leads to the best test performance on MiniGrid. The top row of Figure~\ref{fig:minigrid-results} summarizes these results.

To test our hypothesis, we bin each level into its corresponding difficulty, expressed as ascending, discrete values (note that \algoabbrev{} does not have access to this privileged information). In the bottom row of Figure~\ref{fig:minigrid-results}, we see how the expected difficulty of levels sampled using \algoabbrev{} changes during training for each environment. We observe that as $P_{\text{replay}}$ is updated, levels become sampled according to an implicit curriculum over the training levels that prioritizes progressively harder levels. 
Of particular note, \algoabbrev{} seems to struggle to discover a useful curriculum for around the first $4{,}000$ updates on ObstructedMazeGamut-Medium, at which point it discovers a curriculum that gradually assigns more weight to harder levels. This curriculum enables \algoabbrev{} with access to only $6{,}000$ training levels to attain even higher mean test returns than the uniform-sampling baseline with access to the full set of training levels, of which there are roughly $4$ billion (so our training levels constitute $0.00015\%$ of the total number).

We further tested an extended version of \algoabbrev{} that trains on the full level distribution on these two environments by tracking a buffer of levels with the highest estimated learning potential. We find it outperforms uniform sampling with access to the full level distribution. These additional results are presented in Appendix \ref{section:training_on_full_dist}.

\section{Related Work}
\label{section:related_work}
Several methods for improving generalization in deep RL adapt techniques from supervised learning, including stochastic regularization \citep{igl2019generalization, cobbe2019procgen}, data augmentation \citep{kostrikov2020image, raileanu2021automatic, wang2020mixreg}, and feature distillation \citep{igl2020impact, cobbe2020phasic}. In contrast, \algoabbrev{} modifies only how the next training level is sampled, thereby easily combining with any model or RL algorithm. 

The selective-sampling performed by \algoabbrev{} makes it a form of active learning \citep{cohn1994improving, settles2009active}. Our work also echoes ideas from \citet{DBLP:conf/icml/GravesBMMK17}, who train a multi-armed bandit to choose the next task in multi-task supervised learning, so to maximize gradient-based progress signals. \citet{DBLP:conf/iclr/SharmaJHR18} extend these ideas to multi-task RL, but add the additional requirement of knowing a maximum target return for each task a priori. 
\citet{zhang2020automatic} use an ensemble of value functions for selective goal sampling in the off-policy continuous control setting, requiring prior knowledge of the environment structure to generate candidate goals. 
Unlike \algoabbrev{}, these methods assume the ability to sample tasks or levels based on their structural properties, an assumption that does not typically hold for PCG simulators. Instead, our method automatically uncovers similarly difficult levels, giving rise to a curriculum without prior knowledge of the environment. 


A recent theme in the PCG setting explores adaptively generating levels to facilitate learning \citep{sukhbaatar2017intrinsic, DBLP:conf/gecco/WangLCS19, wang2020enhanced, khalifa2020pcgrl, akkaya2019rubiks, campero2020learning, dennis2020emergent}. Unlike these approaches, our method does not assume control over level generation, requiring only the ability to replay previously visited levels. These methods also require parameterizing level generation with additional learning modules. In contrast, our approach does not require such extensions of the environment, for example including teacher-specific action spaces in the case of \citet{campero2020learning}. Most similar to our method, \citet{Matiisen_2020} proposes a teacher-student curriculum learning (TSCL) algorithm that samples training levels by considering the change in episodic returns per level, though they neither design nor test the method for generalization. As shown in Section \ref{subsection:procgen_benchmark}, it provides inconsistent benefits at test time. Further, unlike TSCL, \algoabbrev{} does not assume access to all levels at the start of training, and as we show in Appendix \ref{section:training_on_full_dist}, \algoabbrev{} can be extended to improve sample efficiency and generalization by training on an unbounded number of training levels.

Like our method, \citet{schaul2015prioritized} and \citet{kapturowski2019r2d2} use TD-errors to estimate learning potential. While these methods make use of TD-errors to prioritize learning from  \emph{past} experiences, our method uses such estimates to prioritize revisiting levels for generating entirely new \emph{future} experiences for learning.

Generalization requires sufficient exploration of environment states and dynamics. Thus, recent exploration strategies \citep[e.g.][]{raileanu2020ride, campero2020learning, zhang2020bebold, zha2021rank} shown to benefit simple PCG settings are complementary to the aims of this work. However, as these studies focus on PCG environments with low-dimensional state spaces,  whether such methods can be successfully applied to more complex PCG environments like Procgen Benchmark remains to be seen. If so, they may potentially combine with \algoabbrev{} to yield additive improvements. We believe the interplay between such exploration methods and \algoabbrev{} to be a promising direction for future research.
\section{Conclusion and Future Work}
We introduced \algoname{} (\algoabbrev{}), an algorithm for selectively sampling the next training level in PCG environments based on the estimated learning potential of revisiting each level for the current policy. We showed that our method remarkably improves both the sample efficiency and generalization of deep RL agents in PCG environments, including the majority of environments in Procgen Benchmark and two challenging MiniGrid environments. We further combined \algoabbrev{} with the prior leading method to set a new state-of-the-art on Procgen Benchmark. Further, on MiniGrid environments, we showed \algoabbrev{} induces an emergent curriculum of increasingly more difficult levels.

The flexibility of the PCG abstraction makes \algoabbrev{} applicable to many problems of practical importance, for example, robotic object manipulation tasks, where domain randomized environment instances map to the notion of levels. We believe \algoabbrev{} may even be applicable to singleton environments, given a procedure for generating variations of the underlying MDP as a function of a level identifier, for example, by varying the starting positions of entities. Another natural extension of PLR is to adapt the method to operate in the goal-conditioned setting, by incorporating goals into the level parameterization. 

Despite the wide applicability of PCG and consequently PLR, not all problem domains can be effectively represented in seed-based simulation. Many real world problems require transfer into domains too complex to be adequately captured by simulation, such as car driving, where realizing a completely faithful simulation would entail solving the very same control problem of interest, creating a chicken-and-egg dilemma. Further, environment resets are not universally available, such as in the continual learning setting, where the agent interacts with the environment without explicit episode boundaries---arguably, a more realistic interaction model for a learning agent deployed in the wild. 

Still, pre-training in simulation with resets can nevertheless benefit such settings, where the target domain is rife with open-ended complexity and where resets are unavailable, especially as training through real-world interactions can be slow, expensive, and precarious. In fact, in practice, we almost exclusively train deep RL policies in simulation for these reasons. As PLR provides a simple method to more fully exploit the simulator for improved test-time performance, we believe PLR can also be adapted to improve learning in these settings.

We further note that while we empirically demonstrated that the L1 value loss acts as a highly effective scoring function, there likely exist even more potent choices. Directly learning such functions may reveal even better alternatives. Lastly, combining \algoabbrev{} with various exploration strategies may further improve test performance in hard exploration environments. We look forward to investigating each of these promising directions in future work, prioritized accordingly, by learning potential. 

\section*{Acknowledgements}
We thank Roberta Raileanu, Heinrich K\"{u}ttler, and Jakob Foerster for useful discussions and feedback on this work, and our anonymous reviewers, for their recommendations on improving this paper.

\sloppy
\bibliography{bib/plrbiblio}
\bibliographystyle{include/icml2021}

\clearpage
\appendix

\section{Experiment Details and Hyperparameters}
\label{app:environments}
\subsection{Procgen Benchmark}
Procgen Benchmark consists of 16 PCG environments of varying styles, exhibiting a diversity of gameplay similar to that of the ALE benchmark. Game levels are determined by a random seed and can vary in navigational layout, visual appearance, and starting positions of entities. All Procgen environments share the same discrete $15$-dimensional action space and produce $64 \times 64 \times 3$ RGB observations. \cite{cobbe2019procgen} provides a comprehensive description of each of the 16 environments. State-of-the-art RL algorithms, like PPO, lead to significant generalization gaps between test and train performance in all games, making Procgen a useful benchmark for assessing generalization performance. 

We follow the standard protocol for testing generalization performance on Procgen: We train an agent for each game on a finite number of levels, $N_{\text{train}}$, and sample test levels from the full distribution of levels. Normalized test returns are computed as $(R - R_{\text{min}})/(R_{\text{max}} - R_{\text{min}})$, where $R$ is the unnormalized return and each game's minimum return, $R_{\text{min}}$, and maximum return, $R_{\text{max}}$, are provided in \citet{cobbe2019procgen}, which uses this same normalization.

To make the most efficient use of our computational resources, we perform hyperparameter sweeps on the easy setting. This also makes our results directly comparable to most prior works benchmarked on Procgen, which have likewise focused on the easy setting. In Procgen easy, our experiments use the recommended settings of $N_{\text{train}} = 200$ and 25M steps of training, as well as the same ResNet policy architecture and PPO hyperparameters shared across all games as in \citet{cobbe2019procgen} and \citet{raileanu2021automatic}. We find 25M steps to be sufficient for uncovering differences in generalization performance among our methods and standard baselines. Moreover, under this setup, we find Procgen training runs require much less wall-clock time than training runs on the two MiniGrid environments of interest over an equivalent number of steps needed to uncover differences in generalization performance. Therefore we survey the empirical differences across various settings of \algoabbrev{} on Procgen easy rather than MiniGrid. 

To find the best hyperparameters for \algoabbrev{}, we evaluate each combination of the scoring function choices in Table~\ref{table:scoring_metrics} with both rank and proportional prioritization, performing a coarse grid search for each pair over different settings of the temperature parameter $\beta$ in $\{0.1, 0.5, 1.0, 1.4, 2.0\}$ and the staleness coefficient $\rho$ in $\{0.1, 0.3, 1.0\}$. For each setting, we run 4 trials across all 16 of games of the Procgen Benchmark, evaluating based on mean unnormalized test return across games. In our TD-error-based scoring functions, we set $\gamma$ and $\lambda$ equal to the same respective values used by the GAE in PPO during training. We found PLR offered the most pronounced gains at $\beta=0.1$ and $\rho=0.1$ on Procgen, but these benefits also held for higher values ($\beta=0.5$ and $\rho=0.3$), though to a lesser degree. 

For UCB-DrAC, we make use of the best-reported hyperparameters on the easy setting of Procgen in \citet{raileanu2021automatic}, listed in Table \ref{tab:hyperparams}.

We found the default setting of mixreg's $\alpha=0.2$, as used by \citet{wang2020mixreg} in the hard setting, performs poorly on the easy setting. Instead, we conducted a grid search over $\alpha$ in $\{0.001, 0.005, 0.01, 0.05, 0.1, 0.2, 0.8, 0.2, 0.5, 0.8, 1\}$.
 
Since the TSCL Window algorithm was not previously evaluated on Procgen Benchmark, we perform a grid search over different settings for both Boltzmann and $\epsilon$-greedy variants of the algorithm to determine the best hyperparameter settings for Procgen easy. We searched over window size $K$ in $\{10,100,1000,10000\}$, bandit learning rate $\alpha$ in $\{0.01, 0.1, 0.5, 1.0\}$, random exploration probability $\epsilon$ in $\{0.0, 0.01, 0.1, 0.5\}$ for the $\epsilon$-greedy variant, and temperature $\tau$ in $\{0.1, 0.5, 1.0\}$ for the Boltzmann variant. Additionally, for a fairer comparison to \algoabbrev{} we further evaluated a variant of TSCL Window that, like \algoabbrev{}, incorporates the staleness distribution, by additionally searching over values of the staleness coefficient $\rho$ in $\{0.0, 0.1, 0.3, 0.5\}$, though we ultimately found that TSCL Window performed best without staleness sampling ($\rho = 0$). 

See Table \ref{tab:hyperparams} for a comprehensive overview of the hyperparameters used for PPO, UCB-DrAC, mixreg, and TSCL Window, shared across all Procgen environments to generate our reported results on Procgen easy.

The evaluation protocol on the hard setting entails training on 500 levels over 200M steps \citep{cobbe2019procgen}, making it more compute-intensive than the easy setting. To save on computational resources, we make use of the same hyperparameters found in the easy setting for each method on Procgen hard, with one exception: As our PPO implementation does not use multi-GPU training, we were unable to quadruple our GPU actors as done in \citet{cobbe2019procgen} and \citet{wang2020mixreg}. Instead, we resorted to doubling the number of environments in our single actor to 128, resulting in mini-batch sizes half as large as used in these two prior works. As such, our baseline results on hard are not directly comparable to theirs. We found setting mixreg's $\alpha=0.2$ as done in \citet{wang2020mixreg} led to poor performance under this reduced batch size. We conducted an additional grid search, finding $\alpha=0.01$ to perform best, as on Procgen easy.

\begin{table}[t!]
\caption{Hyperparameters used for training on Procgen Benchmark and MiniGrid environments.}
\label{tab:hyperparams}
\begin{center}
\begin{tabular}{lrr}
\toprule
\textbf{Parameter} & Procgen & MiniGrid \\
\midrule
\emph{PPO} & & \\
$\gamma$ & 0.999 & 0.999 \\
$\lambda_{\text{GAE}}$ & 0.95 & 0.95 \\
PPO rollout length & 256 & 256 \\
PPO epochs & 3 & 4 \\
PPO minibatches per epoch & 8 & 8 \\
PPO clip range & 0.2 & 0.2 \\
PPO number of workers & 64 & 64 \\
Adam learning rate & 5e-4 & 7e-4 \\
Adam $\epsilon$ & 1e-5 & 1e-5 \\
return normalization & yes & yes \\
entropy bonus coefficient & 0.01 & 0.01 \\
value loss coefficient & 0.5 & 0.5 \\

\addlinespace[10pt]
\emph{PLR} & & \\
Prioritization & rank & rank \\
Temperature, $\beta$, $0.1$ & 0.1 & 0.1 \\
Staleness coefficient, $\rho$ & 0.1 & 0.3 \\

\addlinespace[10pt]
\emph{UCB-DrAC} & & \\
Window size, $K$ & 10 & - \\
Regularization coefficient, $\alpha_r$ & 0.1 & - \\
UCB exploration coefficient, $c$ & 0.1 & - \\

\addlinespace[10pt]
\emph{mixreg} & & \\
Beta shape, $\alpha$ & 0.01 & - \\

\addlinespace[10pt]
\emph{TSCL Window} & & \\
Bandit exploration strategy & $\epsilon$-greedy & - \\
Window size, $K$ & 10 & - \\
Bandit learning rate, $\alpha$ & 1.0 & - \\
Exploration probability, $\epsilon$ & 0.5 & - \\

\bottomrule
\vspace{-1cm}
\end{tabular}
\end{center}
\end{table}

\subsection{MiniGrid}
\label{app:minigrid-description}
The MiniGrid suite \cite{gym_minigrid} features a series of highly structured environments of increasing difficulty. Each environment features a task in a grid world setting, and as in Procgen, environment levels are determined by a seed. Harder levels require the agent to perform longer action sequences over a combinatorially-rich set of game entities, on increasingly larger grids. The clear ordering of difficulty over subsets of MiniGrid environments allows us to track the relative difficulty of levels sampled by \algoabbrev{} over the course of training. 

MiniGrid environments share a discrete 7-dimensional action space and produce a 3-channel integer state encoding of the $7 \times 7$ grid immediately including and in front of the agent. However, following the training setup in \citet{igl2019generalization}, we modify the environment to produce an $N \times M \times 3$ encoding of the full grid, where $N$ and $M$ vary according to the maximum grid dimensions of each environment. Full observability makes generalization harder, requiring the agent to generalize across different level layouts in their entirety. 

We evaluate \algoabbrev{} with rank prioritization on two MiniGrid environments whose levels are uniformly distributed across several difficulty settings. Training on levels of varying difficulties helps agents make use of the easier levels as stepping stones to learn useful behaviors that help the agent make progress on harder levels. However, under the uniform-sampling baseline, learning may be inefficient, as the training process does not selectively train the agent on levels of increasing difficulty, leading to wasted training steps when a difficult level is sampled early in training. On the contrary, if \algoabbrev{} scores levels according to the time-averaged L1 value loss of recently experienced level trajectories, the average difficulty of the sampled levels should adapt to the agent's current abilities, following the reasoning outlined in the Value Correction Hypothesis, stated in Section \ref{section:methods}.

As in \citet{igl2019generalization}, we parameterize the agent policy as a 3-layer CNN with 16, 32, and 32 channels, with a final hidden layer of size 64. All kernels are $2 \times 2$ and use a stride of 1. For the ObstructedMazeGamut environments, we increase the number of channels of the final CNN layer to 64. We follow the same high-level generalization evaluation protocol used for Procgen, training the agent on a fixed set of 4000 levels for MultiRoom-N4-Random, 3000 levels for ObstructedMazeGamut-Easy, and 6000 levels for ObstructedMazeGamut-Medium, and testing on the full level distribution. We chose these values for $|\Lambda_{\text{train}}|$ to ensure roughly 1000 training levels of each difficulty setting of each environment. We model our PPO parameters on those used by \citet{igl2019generalization} in their MiniGrid experiments. We performed a grid search to find that \algoabbrev{} with rank prioritization, $\beta = 0.1$, and $\rho = 0.3$ learned most quickly on the MultiRoom environment, and used this setting for all our MiniGrid experiments. Table \ref{tab:hyperparams} summarizes these hyperparameter choices.

The remainder of this section provides more details about the various MiniGrid environments used in this work.

\begin{figure}[htbp]
    \centering
    \includegraphics[width=\linewidth]{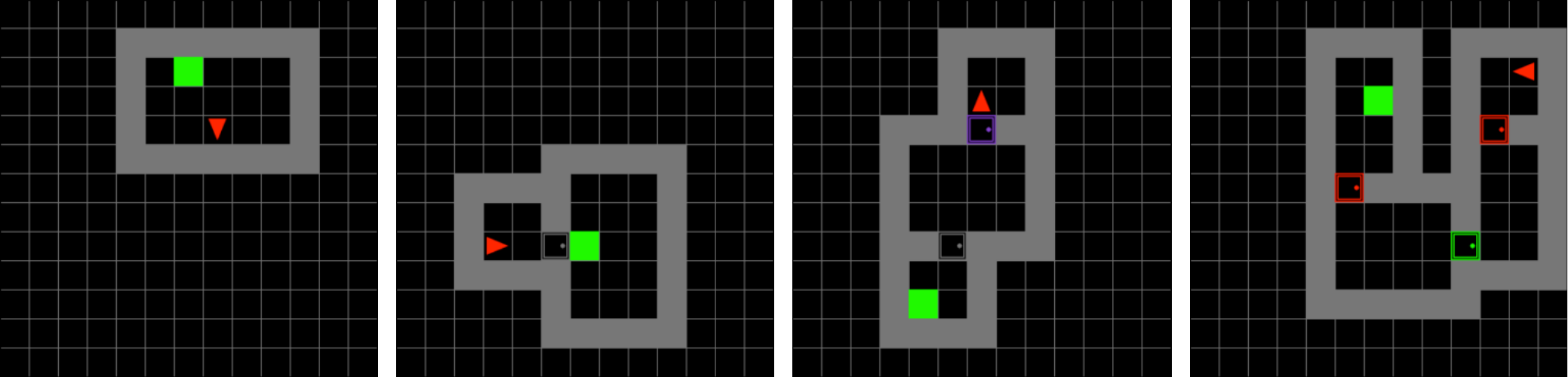}
    \caption{Example levels of each of the four difficulty levels of MultiRoom-N4-Random, in order of increasing difficulty from left to right. The agent (red triangle) must reach the goal (green square).}
    \label{image:mutliroom-n4-random-examples}
\end{figure}

\paragraph{MultiRoom-N4-Random} This environment requires the agent to navigate through 1, 2, 3, or 4 rooms respectively to reach a goal object, resulting in a natural ordering of levels over four levels of difficulty. The agent always starts at a random position in the furthest room from the goal object, facing a random direction. The goal object is also initialized to a random position within its room. See Figure \ref{image:mutliroom-n4-random-examples} for screenshots of example levels.

\begin{figure}[htbp]
    \centering
    \includegraphics[width=1.0\linewidth]{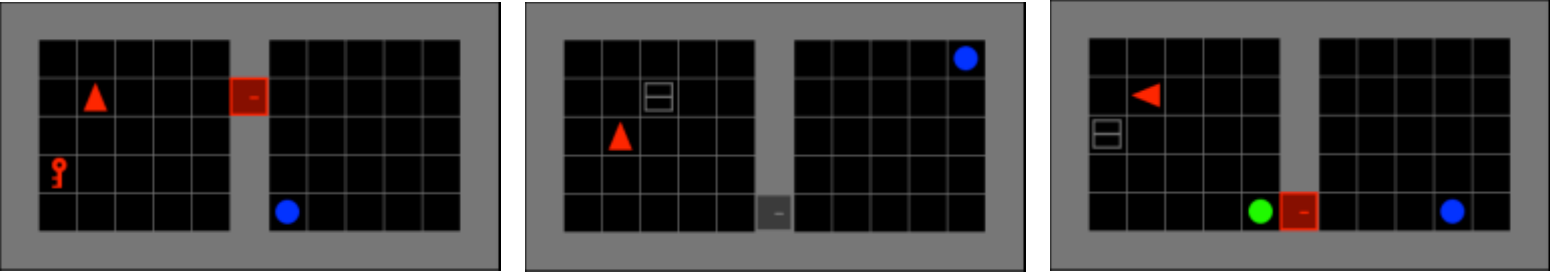}
    \caption{Example levels of each of the three difficulty levels of ObstructedMazeGamut-Easy, in order of increasing difficulty from left to right. The agent must find the key, which may be hidden under a box, to unlock a door, which may be blocked by an obstacle, to reach the goal object (blue circle).}
    \label{image:omg-easy-examples}
\end{figure}

\paragraph{ObstructedMazeGamut-Easy} This environment consists of levels uniformly distributed across the first three difficulty settings of the ObstructedMaze environment, in which the agent must locate and pick up the key in order to unlock the door to pick up a goal object in a second room. The agent and goal object are always initialized in random positions in different rooms separated by the locked door. The second difficulty setting further requires the agent to first uncover the key from under a box before picking up the key. The third difficulty level further requires the agent to first move a ball blocking the door before entering the door. See Figure \ref{image:omg-easy-examples} for screenshots of example levels.

\begin{figure}[htbp]
    \centering
    \includegraphics[width=\linewidth]{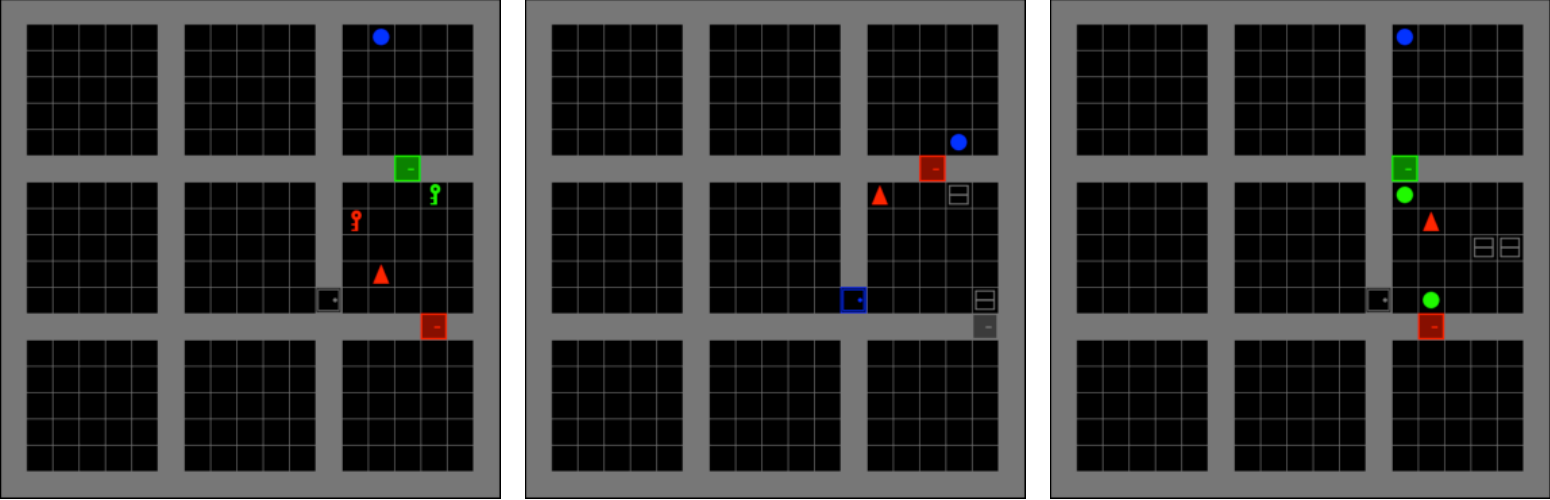}
    \caption{Example levels in increasing difficulty from left to right of each additional difficulty setting introduced by ObstructedMazeGamut-Hard in addition to those in ObstructedMazeGamut-Easy.}
    \label{image:omg-medium-examples}
    \vspace{-0.5cm}
\end{figure}

\paragraph{ObstructedMazeGamut-Hard} This environment consists of levels uniformly distributed across the first six difficulty levels of the ObstructedMaze environment. Harder levels corresponding to the fourth, fifth, and sixth difficulty settings include two additional rooms with no goal object to distract the agent. Each instance of these harder levels also contain two pairs of keys of different colors, each opening a door of the same color. The agent always starts one room away from the randomly positioned goal object. Each of the two keys is visible in the fourth difficulty setting and doors are unobstructed. The fifth difficulty setting hides the keys under boxes, and the sixth again places obstacles that must be removed before entering two of the doors, one of which is always the door to the goal-containing room. See Figure \ref{image:omg-medium-examples} for example screenshots.

\section{Additional Experimental Results}
\label{app:additional_experiments}

\subsection{Extended Results on Procgen Benchmark}
\begin{figure*}[h]
    \centering
    \includegraphics[width=\linewidth]{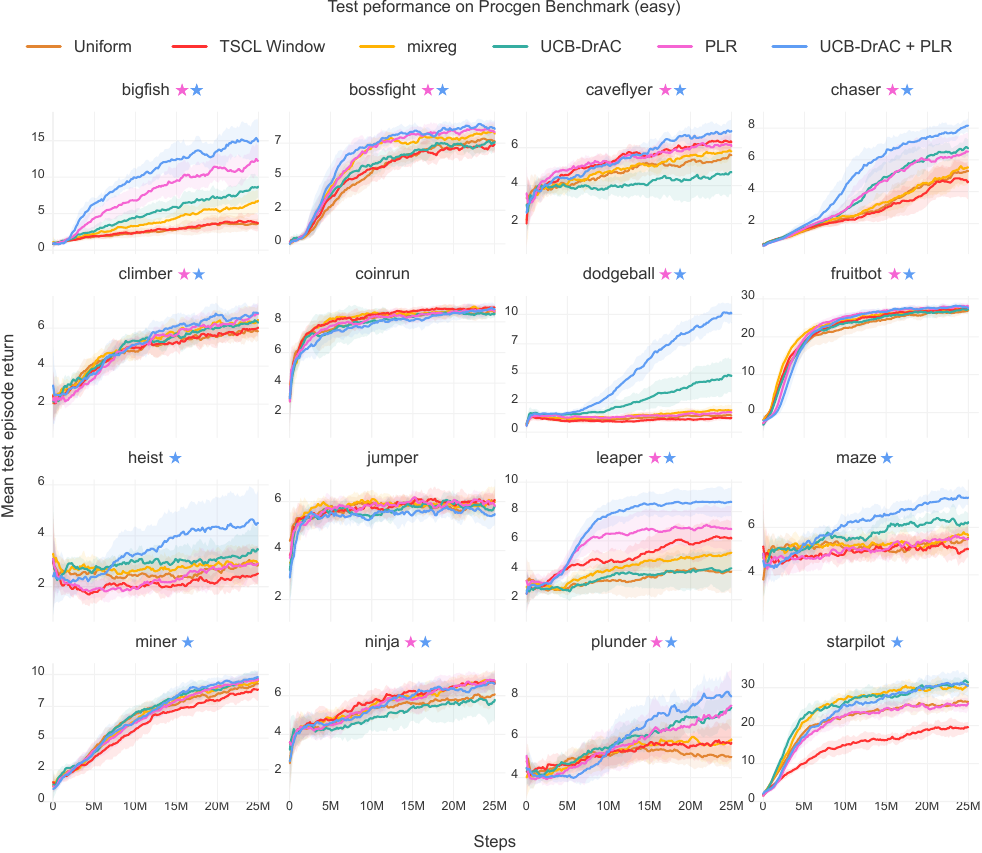}
    \caption{Mean test episode return (10 runs) on each Procgen Benchmark game (easy). The shaded area indicates one standard deviation around the mean. \algoabbrev{}-based methods consistently match or outperform uniform sampling with statistically significance ($p < 0.05$), indicated by a $\bigstar$ of the corresponding color. We see that TSCL results in inconsistent outcomes across games, notably drastically lower test returns on StarPilot.}
    \label{fig:procgen_test_returns_large}
\end{figure*}

\begin{figure*}[h]
    \centering
    \includegraphics[width=\linewidth]{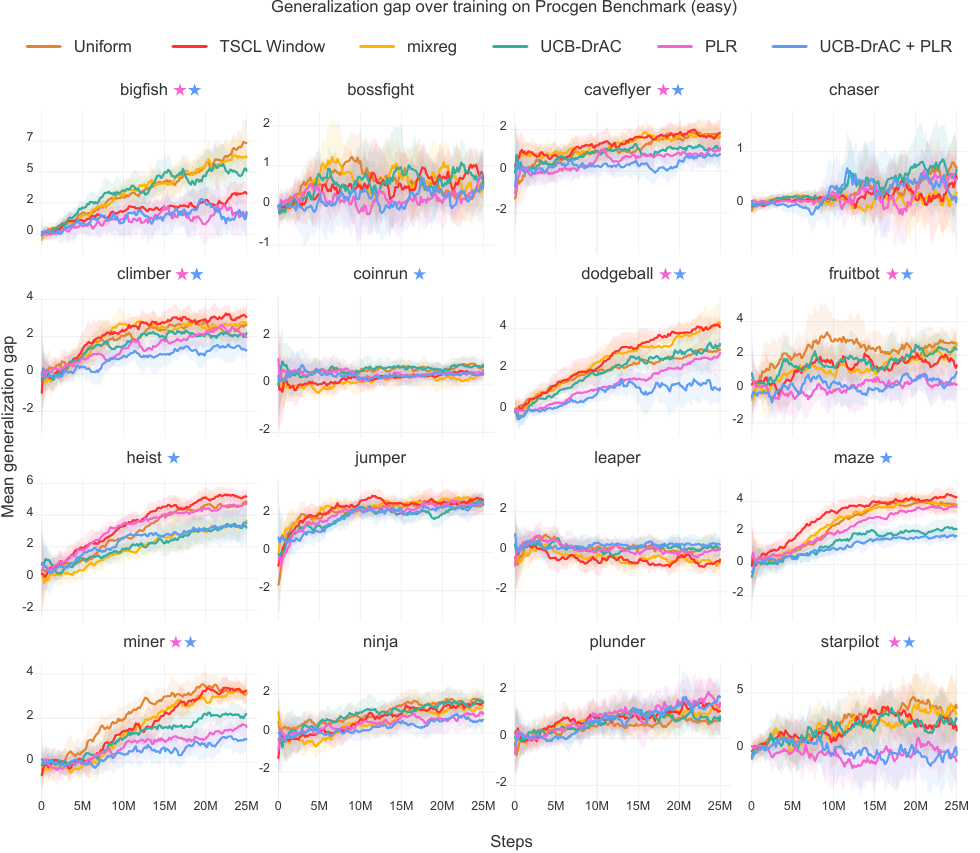}
    \caption{Mean generalization gaps throughout training (10 runs) on each Procgen Benchmark game (easy). The shaded area indicates one standard deviation around the mean. A $\bigstar$ indicates the method of matching color results in a statistically significant ($p < 0.05$) reduction in generalization gap compared to the uniform-sampling baseline. By itself, \algoabbrev{} significantly reduces the generalization gap on 7 games, and UCB-DrAC, on 5 games. This number jumps to 10 of 16 games when these two methods are combined. TSCL only significantly reduces generalization gap on 2 of 16 games relative to uniform sampling, while increasing it on others, most notably on Dodgeball.}
    \label{fig:procgen_gen_gap_large}
\end{figure*}

\begin{figure*}[htbp]
    \centering
    \includegraphics[width=\textwidth]{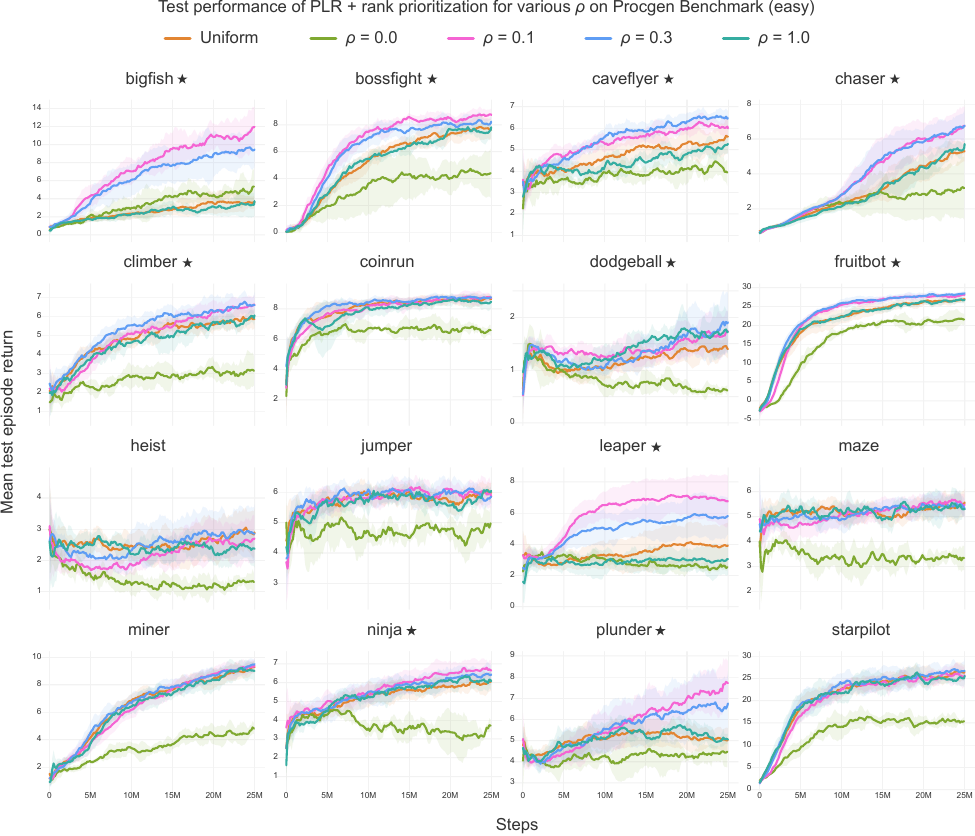}
    \caption{Mean test episode returns (10 runs) on the Procgen Benchmark (easy) for \algoabbrev{} with rank prioritization and $\beta=0.1$ across a range of staleness coefficient values, $\rho$. The replay distribution must consider both the L1 value-loss and staleness values to realize improvements to generalization and sample efficiency. The shaded area indicates one standard deviation around the mean. A $\bigstar$ next to the game name indicates that $\rho = 0.1$ exhibits statistically significantly better final test returns or sample efficiency along the test curve ($p < 0.05$), which we observe in 10 of 16 games.}
    \label{fig:rank_staleness_spectrum}
\end{figure*}

\begin{figure*}
    \centering
    \includegraphics[width=\textwidth]{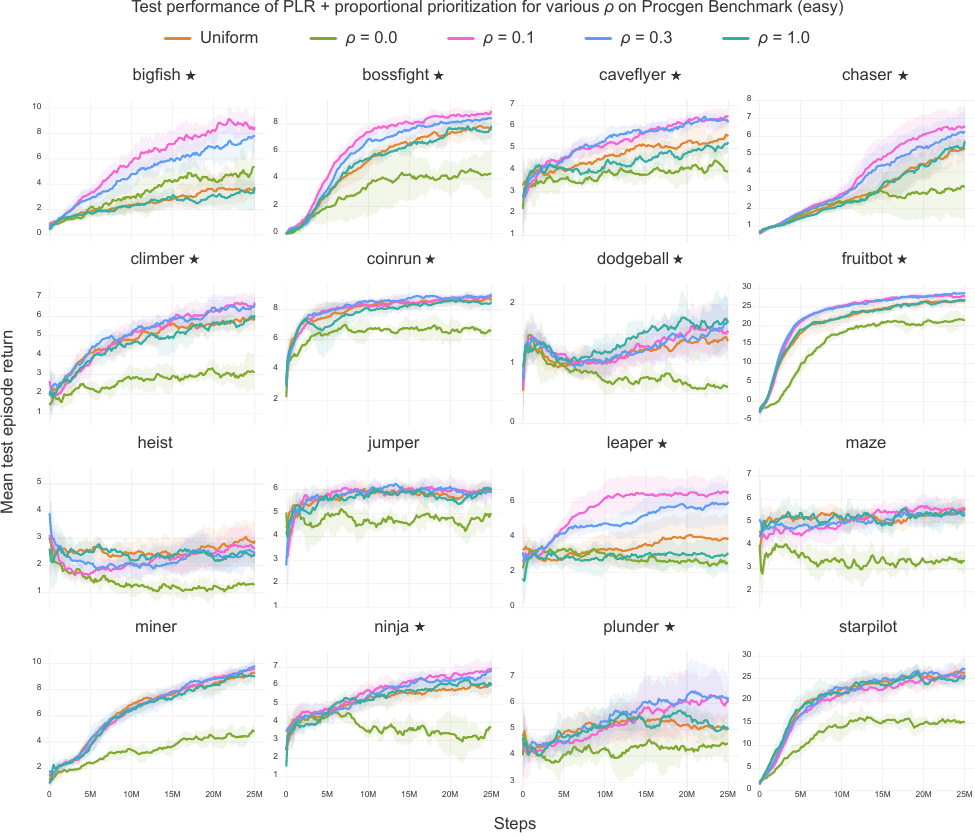}
    \caption{Mean test episode returns (10 runs) on the Procgen Benchmark (easy) for \algoabbrev{} with proportional prioritization and $\beta=0.1$ across a range of values of $\rho$. As in the case of rank prioritization, the replay distribution must consider both the L1 value loss score and staleness values in order to realize performance improvements. The shaded area indicates one standard deviation around the mean. A $\bigstar$ next to the game name indicates the condition $\rho = 0.1$ exhibits statistically significantly better final test returns or sample efficiency along the test curve ($p<0.05$), which we observe in 11 of 16 games.}
    \label{fig:prop_staleness_spectrum}
\end{figure*}

We present an overview of the improvements in test performance of each method across all 16 Procgen Benchmark games over 10 runs in Figure \ref{fig:procgen_test_returns_large}. For each game, Figure \ref{fig:procgen_gen_gap_large} further shows how the  generalization gap changes over the course of training under each method tested. 
We show in Figures~\ref{fig:rank_staleness_spectrum} and~\ref{fig:prop_staleness_spectrum}, the mean test episodic returns on the Procgen Benchmark (easy) for \algoabbrev{} with rank and proportional prioritization, respectively. In both of these plots, we can see that using only staleness ($\rho=1$) or only L1 value loss scores ($\rho=0$) is considerably worse than direct level sampling. Thus, we only observe gains compared to the baseline when both level scores and staleness are used for the sampling distribution. Comparing  Figures~\ref{fig:rank_staleness_spectrum} with~\ref{fig:prop_staleness_spectrum} we find that \algoabbrev{} with proportional instead of rank prioritization provides statistically significant gains over uniform level sampling on an additional game (CoinRun), but rank prioritization leads to slightly larger mean improvements on several games.

\begin{figure*}[htpb]
    \centering
    \includegraphics[width=\textwidth]{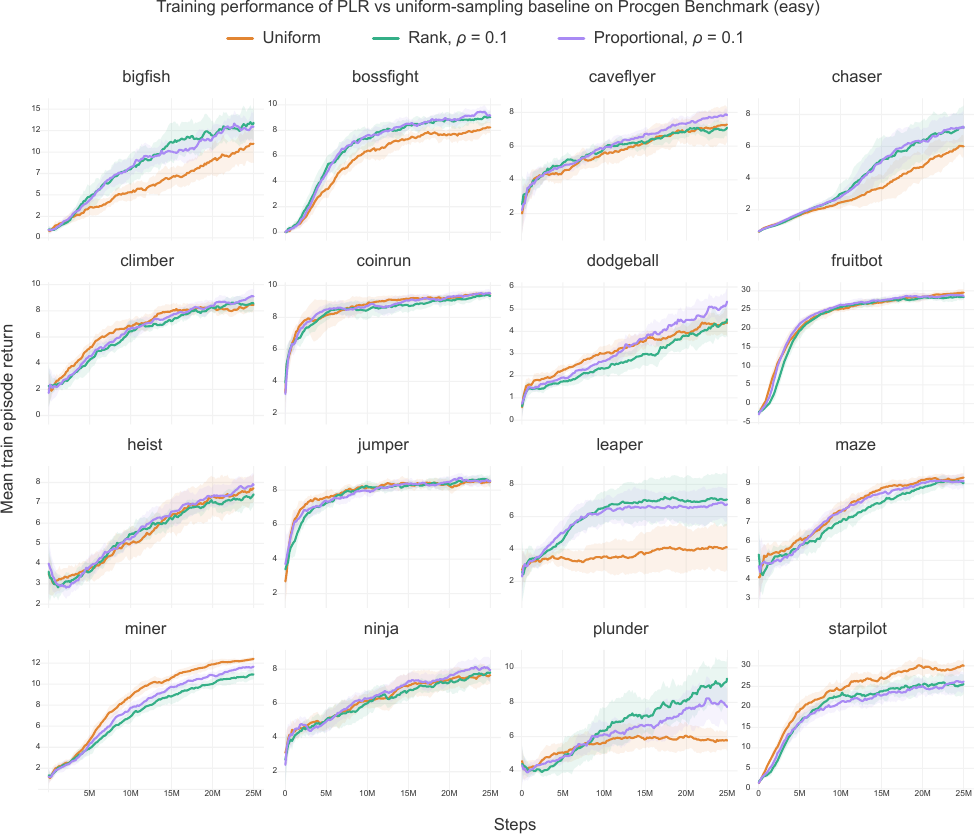}
    \caption{Mean training episode returns (10 runs) on the Procgen Benchmark for (easy) \algoabbrev{} with $\beta=0.1$, $\rho=0.1$, and each of rank and proportional prioritization. On some games, \algoabbrev{} improves both training sample efficiency and generalization performance (e.g. BigFish and Chaser), while on others, only generalization performance (e.g.  CaveFlyer with rank prioritization). The shaded area indicates one standard deviation around the mean.}
    \label{fig:rank_prop_train_eval}
\end{figure*}

Figure~\ref{fig:rank_prop_train_eval} shows that when \algoabbrev{} improves generalization performance, it also either matches or improves training sample efficiency, suggesting that when beneficial to test performance, the representations learned via the auto-curriculum induced by \algoabbrev{} prove similarly useful on training levels. However we see that our method reduces training sample efficiency on two games on which our method does not improve generalization performance. Since our method does not discover useful auto-curricula for these games, it is likely that uniformly sampling levels at training time allows the agent to better memorize useful behaviors on each of the training levels compared to the selective sampling performed by our method.

\begin{figure*}[b!]
    \centering
    \includegraphics[width=\textwidth]{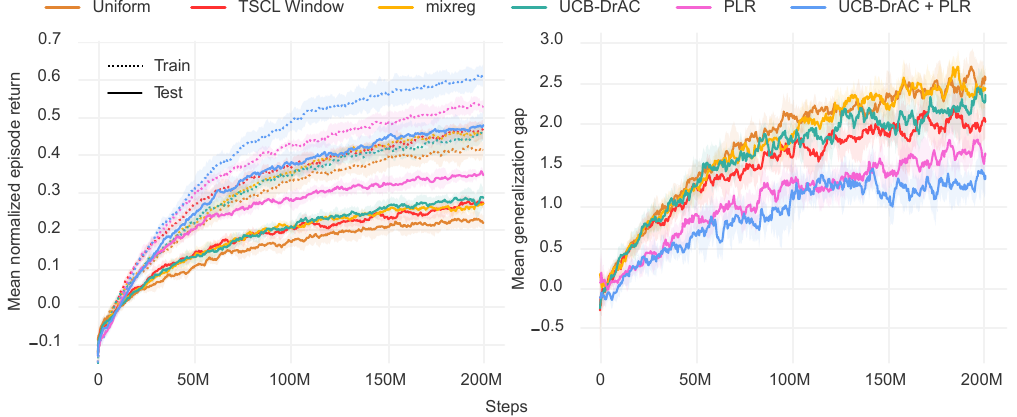}
    \caption{Left: Mean normalized train and test episode returns on Procgen Benchmark (hard). Right: Corresponding generalization gaps during training. All curves are averaged across all environments over 5 runs. The shaded area indicates one standard deviation around the mean. PLR-based methods statistically significantly outperform all others in both train and test returns. Only the PLR-based methods statistically significantly reduce the generalization gap ($p < 0.05$).}
    \label{fig:procgen_hard_summary}
\end{figure*}

\begin{figure*}[!hbtp]
    \centering
    \includegraphics[width=\textwidth]{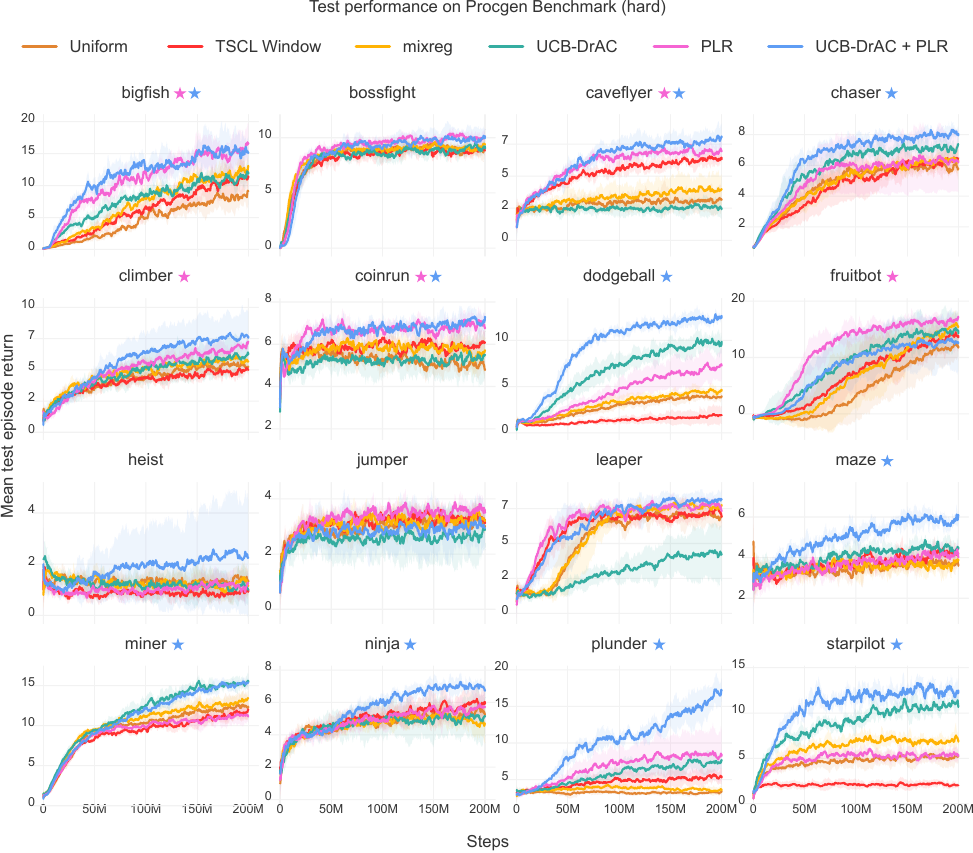}
    \caption{Mean test episode returns (5 runs) on Procgen Benchmark (hard), using best hyperparameters found on the easy setting. The shaded area indicates one standard deviation around the mean. A $\bigstar$ indicates statistically significant improvement over the uniform-sampling baseline by the PLR-based method of the matching color ($p < 0.05$).}
    \label{fig:procgen_hard_test_all}
\end{figure*}

Finally, we also benchmarked PLR and UCB-DrAC + PLR against uniform sampling, TSCL Window, mixreg, and UCB-DrAC on Procgen hard across 5 runs per environment. Due to the high computational cost of the evaluation protocol for Procgen hard, which entails 200M training steps, we directly use the best hyperparameters found in the easy setting for each method. The results in Figure \ref{fig:procgen_hard_summary} show the two PLR-based methods significantly outperform all other methods in terms of normalized mean train and test episodic return, as well as reduction in mean generalization gap, attaining even greater margins of improvement than in the easy setting. As summarized by Table \ref{tab:ucb-drac-plr-hard}, the gains of PLR and UCB + PLR in mean normalized test return relative to uniform sampling in the hard setting are comparable to those in the easy setting. We provide plots of episodic test return over training for each individual environment in Figure \ref{fig:procgen_hard_test_all}.


\subsection{Extended Results on Minigrid}
\begin{figure}[t!]
\centering
\includegraphics[width=\linewidth]{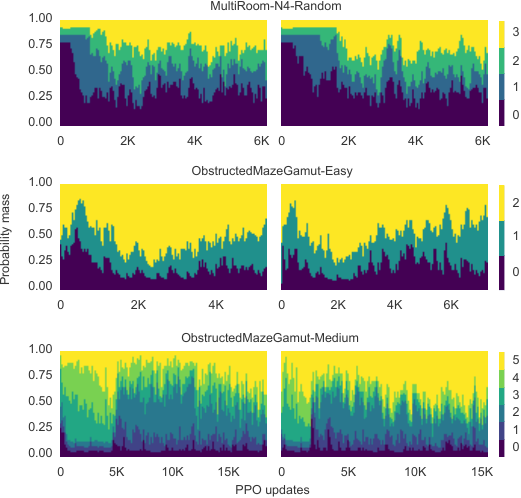}
\caption{PLR consistently induces emergent curricula from easier to harder levels during training. Left and right correspond to two additional training runs independent of that in Figure \ref{fig:minigrid-results}.} 
\label{fig:minigrid-curriculum-extra}
\end{figure}

To demonstrate that PLR consistently induces an emergent curriculum, we present plots showing the change in probability mass over different difficulty bins for additional training runs in Figure \ref{fig:minigrid-curriculum-extra}. Like in Figure \ref{fig:minigrid-results}, we see the probability mass assigned by $P_{\textnormal{replay}}$ gradually shifts from easier to harder levels over the course of training.

\subsection{Training on the Full Level Distribution}
\label{section:training_on_full_dist}

\begin{figure}[t!]
\centering
\begin{subfigure}[b]{1.0\linewidth}
    \includegraphics[width=\linewidth]{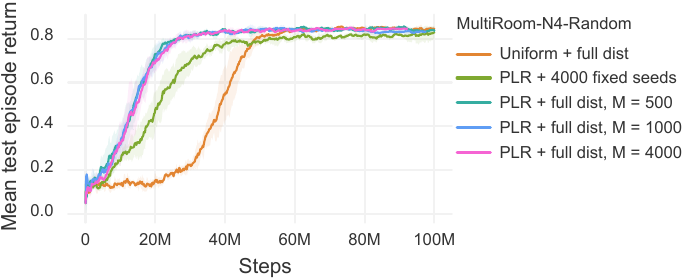}
\end{subfigure}

\begin{subfigure}[b]{1.0\linewidth}
    \includegraphics[width=\linewidth]{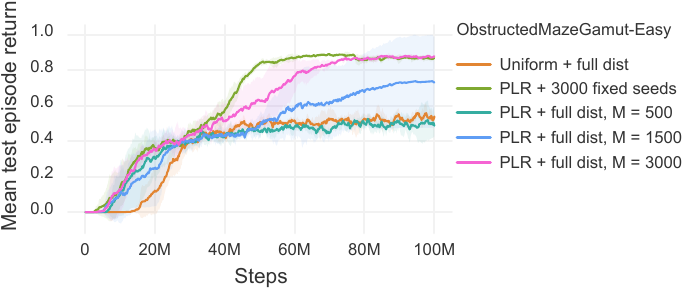}
\end{subfigure}

\caption{Mean test episodic returns on MultiRoom-N4-Random (top) and ObstructedMazeGamut-Easy (bottom) with access to the full level distribution at training. Plots are averaged over 3 runs. We set $P_{D}$ to a Bernoulli parameterized as $p = 0.5$ for MultiRoom-N4-Random and $p=0.95$ for ObstructedMazeGamut-Easy (found via grid search). As with all MiniGrid experiments using \algoabbrev{}, we use rank prioritization, $\beta = 0.1$, and $\rho = 0.3$.} 

\label{fig:minigrid-full-dist}
\end{figure}

While assessing generalization performance calls for using a fixed set of training levels, ideally our method can also make use of the full level distribution if given access to it. We take advantage of an unbounded number of training levels by modifying the list structures for storing scores and timestamps (see Algorithm \ref{alg:pg_level_replay} and \ref{alg:level_replay_mc}) to track the top $M$ levels by learning potential in our finite level buffer. When the lists are full, we set the next level for replacement to be 
\[
l_{\text{min}} = \underset{l}{\arg \min}\;P_{\text{replay}}(l).
\]
When the outcome of the Bernoulli $P_{D}$ entails sampling a new level $l$, the score and timestamps of $l$ replace those of $l_{\text{min}}$ only if the score of $l_{\text{min}}$ is lower than that of $l$. In this way, \algoabbrev{} keeps a running buffer throughout training of the top $M$ levels appraised to have the highest learning potential for replaying anew.

Figure \ref{fig:minigrid-full-dist} shows that with access to the full level distribution at training, \algoabbrev{} improves sample efficiency and generalization performance in both environments compared to uniform sampling on the full distribution. In MultiRoom-N4-Random, the value $M$ makes little difference to test performance, and training with \algoabbrev{} on the full level distribution leads to a policy outperforming one trained with \algoabbrev{} on a fixed set of training levels. However, on ObstructedMazeGamut-Easy, a smaller $M$ leads to worse test performance. Nevertheless, for all but $M=500$, including the case of a fixed set of 3000 training levels, \algoabbrev{} leads to better mean test performance than uniform sampling on the full level distribution.

\section{Algorithms}
\label{app:algorithms}

In this section, we provide detailed pseudocode for how \algoabbrev{} can be used for experience collection when using $T$-step rollouts. Algorithm~\ref{alg:t-step_pg_level_replay} presents the extension of the generic policy-gradient training loop presented in Algorithm~\ref{alg:pg_level_replay} to the case of $T$-step rollouts, and Algorithm~\ref{alg:t-step_collect_exp_prioritized_level_replay} presents an implementation of experience collection in this setting (extending Algorithm~\ref{alg:level_replay_mc}). Note that when using $T$-step rollouts in the training loop, rollouts may start and end between episode boundaries. 
To compute level scores on full trajectories segmented across rollouts, we compute scores of partial episodes according to Equation~\ref{eq:average_scoring_function}, and record these partial scores alongside the partial episode step count in a separate buffer $\Tilde{S}$. The function $\textbf{score}$ then technically, optionally takes the additional input $\Tilde{S}$ (through an abuse of notation) as an additional argument to stitch together this partial information into scores of full episodic trajectories.


\begin{table*}[t!]
\setlength{\tabcolsep}{2.5pt}
\caption{Comparison of test scores of PPO with \algoabbrev{} against PPO with uniform-sampling on the hard setting of Procgen Benchmark. Following \citep{raileanu2021automatic}, reported figures represent the mean and standard deviation of average test scores over 100 episodes aggregated across 5 runs, each initialized with a unique training seed. For each run, a normalized average return is computed by dividing the average test return for each game by the corresponding average test return of the uniform-sampling baseline over all 500 test episodes of that game, followed by averaging these normalized returns over all 16 games. The final row reports the mean and standard deviation of the normalized returns aggregated across runs. Bolded methods are not significantly different from the method with highest mean, unless all are, in which case none are bolded.}
\label{tab:ucb-drac-plr-hard}
\begin{center}
\begin{tabular}{l|r r r r | r r}
\toprule
Environment &Uniform &TSCL &mixreg &UCB-DrAC &PLR &UCB-DrAC + PLR\\
\midrule
BigFish&$9.7\pm1.8$&$\mathbf{11.9\pm2.5}$&$\mathbf{12.0\pm2.5}$&$10.9\pm1.6$&$\mathbf{15.3\pm3.6}$&$\mathbf{15.5\pm2.8}$\\
BossFight&$\mathbf{9.6\pm0.2}$&$8.4\pm0.7$&$\mathbf{9.3\pm0.9}$&$9.0\pm0.2$&$\mathbf{9.7\pm0.4}$&$\mathbf{9.5\pm1.1}$\\
CaveFlyer&$3.5\pm0.8$&$6.3\pm0.6$&$4.0\pm1.0$&$2.6\pm0.8$&$6.4\pm0.6$&$\mathbf{8.0\pm0.9}$\\
Chaser&$5.9\pm0.5$&$6.2\pm1.0$&$\mathbf{6.5\pm0.8}$&$\mathbf{7.0\pm0.6}$&$\mathbf{6.8\pm2.2}$&$\mathbf{7.6\pm0.2}$\\
Climber&$ 5.3\pm1.1$&$ 5.2\pm0.7$&$ 5.7\pm0.7$&$ 6.1\pm1.0$&$ 7.4\pm0.6$&$ 7.6\pm1.8$\\
CoinRun&$4.5\pm0.4$&$5.8\pm0.8$&$\mathbf{6.2\pm1.0}$&$5.2\pm1.0$&$\mathbf{6.8\pm0.6}$&$\mathbf{7.1\pm0.5}$\\
Dodgeball&$3.9\pm0.6$&$1.9\pm0.9$&$4.7\pm1.0$&$9.9\pm1.2$&$7.4\pm1.3$&$\mathbf{12.4\pm0.7}$\\
FruitBot&$\mathbf{11.9\pm4.2}$&$13.1\pm2.3$&$\mathbf{14.7\pm2.2}$&$\mathbf{15.6\pm3.7}$&$\mathbf{16.7\pm1.0}$&$\mathbf{12.9\pm5.1}$\\
Heist&$1.5\pm0.4$&$0.9\pm0.3$&$1.2\pm0.4$&$1.1\pm0.3$&$1.3\pm0.4$&$2.6\pm2.2$\\
Jumper&$3.2\pm0.3$&$3.2\pm0.3$&$3.3\pm0.4$&$2.9\pm0.9$&$3.5\pm0.5$&$3.3\pm0.8$\\
Leaper&$7.1\pm0.3$&$\mathbf{7.5\pm0.5}$&$\mathbf{7.5\pm0.5}$&$3.8\pm1.6$&$\mathbf{7.4\pm0.2}$&$\mathbf{8.2\pm0.7}$\\
Maze&$3.6\pm0.7$&$3.8\pm0.6$&$3.9\pm0.5$&$4.4\pm0.2$&$4.0\pm0.4$&$\mathbf{6.2\pm0.4}$\\
Miner&$12.8\pm1.4$&$11.7\pm0.9$&$13.3\pm1.6$&$\mathbf{16.1\pm0.6}$&$11.3\pm0.7$&$\mathbf{15.3\pm0.8}$\\
Ninja&$5.2\pm0.1$&$5.9\pm0.8$&$5.0\pm1.0$&$5.2\pm1.0$&$\mathbf{6.1\pm0.6}$&$\mathbf{6.9\pm0.3}$\\
Plunder&$3.2\pm0.1$&$5.4\pm1.1$&$3.7\pm0.4$&$7.8\pm1.1$&$8.6\pm2.7$&$\mathbf{17.5\pm1.3}$\\
StarPilot&$5.5\pm0.6$&$2.1\pm0.4$&$6.9\pm0.6$&$\mathbf{11.2\pm1.7}$&$5.4\pm0.8$&$\mathbf{12.3\pm1.5}$\\
\midrule
Normalized test returns (\%)&$100.0\pm2.0$&$103.9\pm3.5$&$110.6\pm3.9$&$126.6\pm3.0$&$135.0\pm6.1$&$\mathbf{182.9\pm8.2}$\\
\bottomrule
\end{tabular}
\end{center}
\end{table*}

\begin{figure*}[!hbtp]
    \centering
    \includegraphics[width=\textwidth]{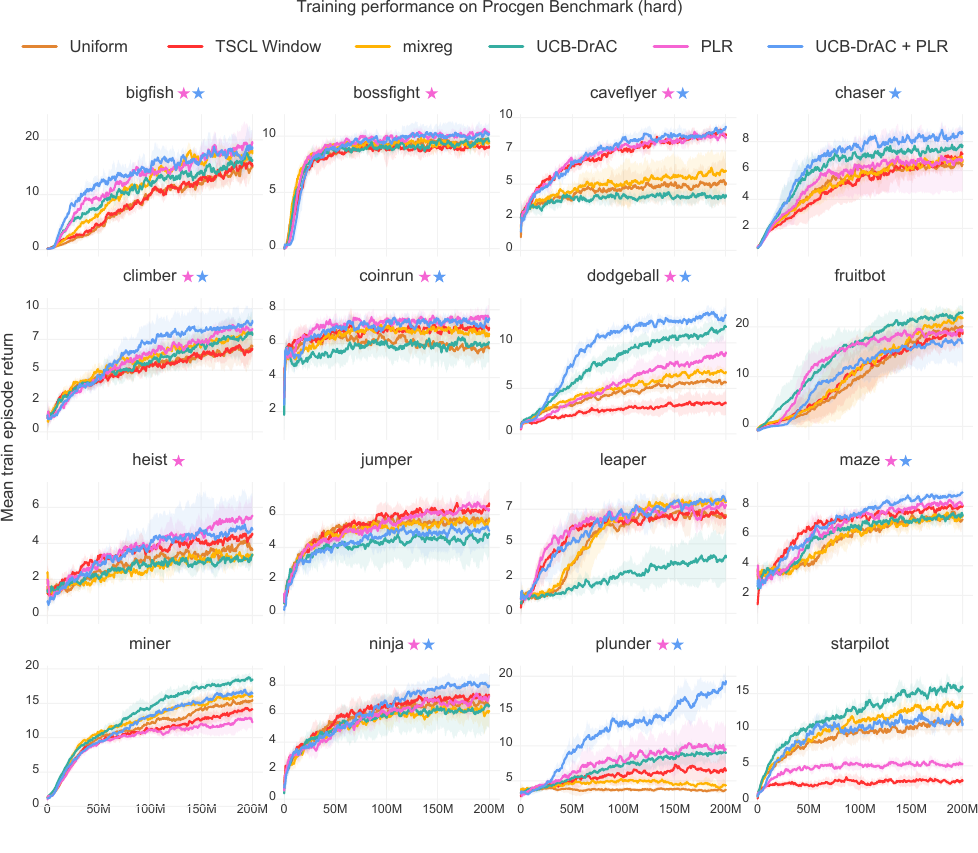}
    \caption{Mean train episode returns (5 runs) on Procgen Benchmark (hard), using the best hyperparameters found on the easy setting. The shaded area indicates one standard deviation around the mean. A $\bigstar$ indicates statistically significant improvement over the uniform-sampling baseline by the PLR-based method of the matching color ($p < 0.05$).  Note that while PLR reduces training performance on StarPilot, it performs comparably to the uniform-sampling baseline at test time, indicating less overfitting to training levels.}
    \label{fig:procgen_hard_train_all}
\end{figure*}

\begin{figure*}[!hbtp]
    \centering
    \includegraphics[width=\textwidth]{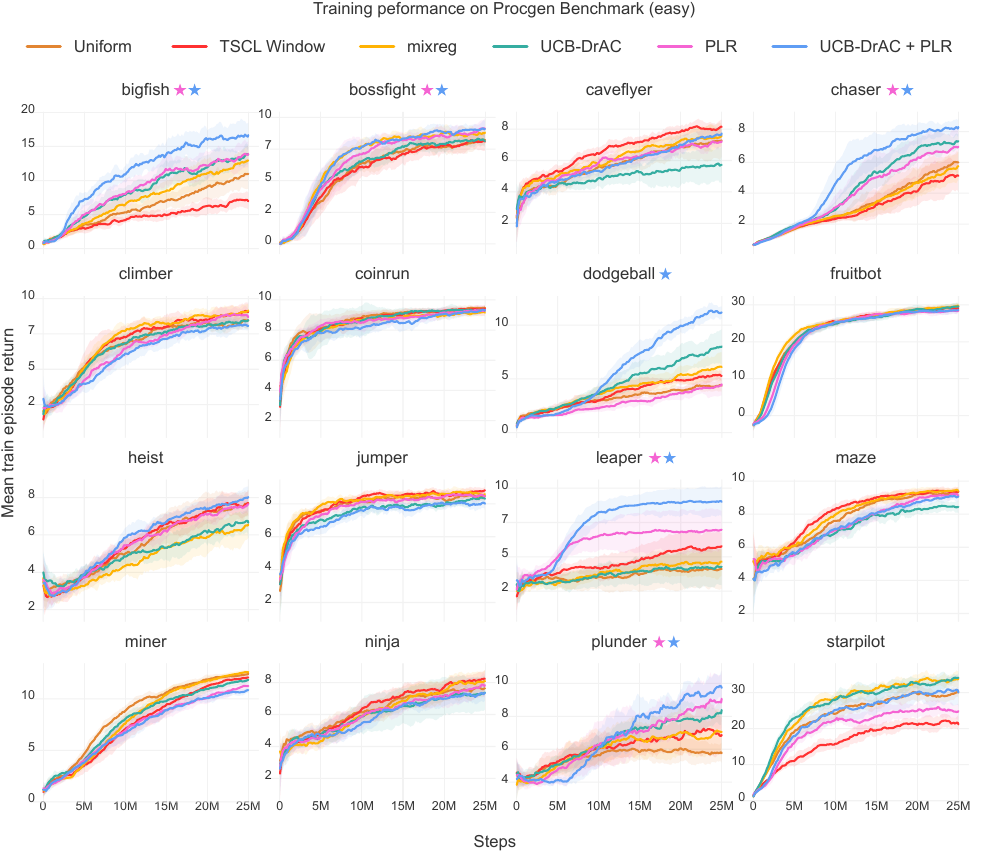}
    \caption{Mean train episode returns (5 runs) on Procgen Benchmark (easy). The shaded area indicates one standard deviation around the mean. A $\bigstar$ indicates statistically significant improvement over the uniform-sampling baseline by the PLR-based method of the matching color ($p < 0.05$). PLR tends to improve or match training sample efficiency. Note that while PLR reduces training performance on StarPilot, it performs comparably to the uniform-sampling baseline at test time, indicating less overfitting to training levels.}
    \label{fig:procgen_easy_train_all}
\end{figure*}

\clearpage

\begin{algorithm*}[t!]
\caption{Generic $T$-step policy-gradient training loop with prioritized level replay}
\label{alg:t-step_pg_level_replay}
\begin{algorithmic}
    \REQUIRE Training levels $\Lambda_{\text{train}}$ of an environment, policy $\pi_{\theta}$, rollout length $T$, number of updates $N_{u}$, batch size $N_{b}$, \\ policy update function $\mathcal{U}(\mathcal{B}, \theta) \rightarrow \theta'$.
    \STATE Initialize level scores $S$, partial level scores $\Tilde{S}$, and level timestamps $C$ 
    \STATE Initialize global episode count $c \gets 0$
    \STATE Initialize set of visited levels $\Lambda_{\text{seen}} = \varnothing$ 
    \STATE Initialize experience buffer $\mathcal{B} = \varnothing$
    \STATE Initialize $N_b$ parallel environment instances $E$, each set to a random level in $\in \Lambda_{\text{train}}$
    \FOR{$u=1$ \textbf{to} $N_{u}$}
        \STATE $\mathcal{B} \leftarrow \varnothing$
        \FOR{$k=1$ \textbf{to} $N_{b}$}
            \STATE $\mathcal{B} \leftarrow \mathcal{B}\;\cup\;\textbf{collect\_experiences}(k, E, \Lambda_{\text{train}}, \Lambda_{\text{seen}}, \pi_{\theta}, T, S, \Tilde{S}, C, c)$ \LineComment{Using Algorithm~\ref{alg:t-step_collect_exp_prioritized_level_replay}}
        \ENDFOR
        \STATE $\theta \leftarrow \mathcal{U}(\mathcal{B}, \theta)$
    \ENDFOR
\end{algorithmic}
\end{algorithm*}

\begin{algorithm*}[htbp]
\caption{Collect $T$-step rollouts with prioritized level replay}
\label{alg:t-step_collect_exp_prioritized_level_replay}
\begin{algorithmic}
    \REQUIRE Actor index $k$, batch environments $E$, training levels $\Lambda_{\text{train}}$, visited levels $\Lambda_{\text{seen}}$, current level $l$, policy $\pi_{\theta}$, rollout length $T$, scoring function $\mathbf{score}$, level scores $S$, partial scores $\Tilde{S}$, staleness values $C$, and global episode count $c$.
    \ENSURE Experience buffer $\mathcal{B}$
    
    \STATE Initialize $\mathcal{B} = \varnothing$, and set current level $l_i = E_k$ 
    \STATE Observe current state $s_0$, termination flag $d_0$
    \IF{$d_0$}
        \STATE Define new index $i \gets |S| + 1$
        \STATE Choose current level $l_i \gets$ $\mathbf{sampleNextLevel}(\Lambda_{\text{train}}, S, C, c)$ and $E_k \gets l_i$
        \STATE Update level timestamp $C_i \gets c$
        \STATE Observe initial state $s_0$
    \ENDIF
    \STATE Choose $a_0 \sim \pi_{\theta}(\cdot|s_0)$
    \STATE t = 1
    \STATE Initialize episodic trajectory buffer $\tau = \varnothing$
    \WHILE{$t < T$}
        \STATE Observe $(s_t, r_t, d_t)$
        \STATE $\mathcal{B}\leftarrow \mathcal{B}\;\cup (s_{t-1}, a_{t-1}, s_t, r_t, d_t, \log\pi_{\theta}(a))$
        \STATE $\tau\leftarrow \tau \cup (s_{t-1}, a_{t-1}, s_t, r_t, d_t, \log\pi_{\theta}(a))$
        \IF{$d_t$}
            \STATE Update level score $S_i \gets \mathbf{score}(\tau, \pi_{\theta},  \Tilde{S}_i)$ and partial score $\Tilde{S}_i \gets 0$
            \STATE $\tau \leftarrow \varnothing$
            \STATE Define new index $i \gets |S| + 1$
            \STATE Update current level $l_i \gets$ {$\textbf{sampleNextLevel}(\Lambda_{\text{train}}, S, C, c)$} and $E_k \gets l_i$
            \STATE Update level timestamp $C_i \gets c$
        \ENDIF
        \STATE Choose $a_{t+1} \sim \pi_{\theta}(\cdot|s_t)$
        \STATE $t \leftarrow t + 1$
    \ENDWHILE
    \IF{not $d_t$}
        \STATE $\Tilde{S}_i \gets (\textbf{score}(\tau, \pi_{\theta}), |\tau|)$ \LineComment{Track partial time-averaged score and $|\tau|$}
    \ENDIF
    \STATE
    \FUNCTION{$\text{sampleNextLevel}(\Lambda_{\text{train}}, S, C, c)$}
        \STATE $c \gets c + 1$
        \STATE Sample replay decision $d \sim{} P_D(d)$
        \IF{$d = 0$ \textbf{and} $\left|\Lambda_{\text{train}} \setminus \Lambda_{\text{seen}}\right| > 0$} 
            \STATE Define new index $i \gets |S| + 1$
            \STATE Sample $l_i \sim{} P_{\text{new}}(l | \Lambda_{\text{train}}, \Lambda_{\text{seen}})$
            \LineComment{Sample an unseen level, if any}
            \STATE Add $l_i$ to $\Lambda_{\text{seen}}$, add initial value $S_i = 0$ to $S$ and $C_i = 0$ to $C$
        \ELSE 
            \STATE Sample $l_i \sim{} (1-\rho) \cdot P_S(l | \Lambda_{\text{seen}}, S) + \rho \cdot P_C(l | \Lambda_{\text{seen}}, C, c)$
            \LineComment{Sample a level for replay}
        \ENDIF
        \STATE \textbf{return} $l_i$
    \ENDFUNCTION
\end{algorithmic}
\end{algorithm*}

\end{document}